\documentclass{article}
\usepackage[utf8]{inputenc}

\pdfoutput=1

\usepackage[backend=biber, style=numeric, sorting=nyt]{biblatex}
\usepackage{amsmath}
\usepackage{amssymb}
\usepackage{graphicx}
\usepackage{tabularx}
\usepackage[margin=1in]{geometry}
\usepackage[font={small,it}]{caption}
\usepackage{authblk}
\usepackage{hyperref}
\usepackage{threeparttable}
\usepackage{caption}
\usepackage{subcaption}

\usepackage{tabulary}
\usepackage{booktabs}
\usepackage[table]{xcolor}

\title{\texttt{PDE-LEARN}: Using Deep Learning to Discover Partial Differential Equations from Noisy, Limited Data}
\author[a]{Robert Stephany\thanks{Corresponding Author. 

Email address: \href{mailto:rrs254@cornell.edu}{\texttt{rrs254@cornell.edu}}}$^{,}$}
\author[a,b]{Christopher Earls}
\affil[a]{\it Center for Applied Mathematics, Cornell University, Ithaca, NY 14850, United States}
\affil[b]{\it School of Civil \& Environmental Engineering, Cornell University, Ithaca, NY 14850, United States}
\date{}
\begin{document}

\maketitle

\begin{center} {\bf Abstract} \end{center}
In this paper, we introduce \texttt{PDE-LEARN}, a novel deep learning algorithm that can identify governing partial differential equations (PDEs) directly from noisy, limited measurements of a physical system of interest. 
\texttt{PDE-LEARN} uses a Rational Neural Network, $U$, to approximate the system response function and a sparse, trainable vector, $\xi$, to characterize the hidden PDE that the system response function satisfies. 
Our approach couples the training of $U$ and $\xi$ using a loss function that (1) makes $U$ approximate the system response function, (2) encapsulates the fact that $U$ satisfies a hidden PDE that $\xi$ characterizes, and (3) promotes sparsity in $\xi$ using ideas from iteratively reweighted least-squares. 
Further, \texttt{PDE-LEARN} can simultaneously learn from several data sets, allowing it to incorporate results from multiple experiments.
This approach yields a robust algorithm to discover PDEs directly from realistic scientific data.
We demonstrate the efficacy of \texttt{PDE-LEARN} by identifying several PDEs from noisy and limited measurements.

$~$


\newpage
\section{Introduction}
\label{sec:Introduction}

Scientific progress is contingent upon finding predictive models for the natural world.
Historically, scientists have discovered new laws by studying physical systems to distill the first principles that govern those systems.
This {\it first-principles} approach has yielded predictive models in many fields, including fluid mechanics, population dynamics, quantum mechanics, and general relativity. 
Unfortunately, despite years of concerted effort, many systems currently lack predictive models, particularly those in the biological sciences \cite{abreu2019mortality}, \cite{amor2020transient}.
    
\bigskip

These roadblocks call for a new approach to identifying governing equations.
At its core, discovering new scientific laws relies on identifying simple governing equations from complex data sets.
Pattern recognition is, notably, one of the core goals of machine learning.
Advances in machine learning, specifically deep learning, offer an intriguing alternative approach to discovering scientific laws.
The burgeoning field of physics-informed machine learning seeks to use machine learning for scientific and engineering applications by designing models whose architecture incorporates knowledge of the physical system.
PDE discovery, a sub-field of physics-informed machine learning, seeks to use machine learning models to identify partial differential equations (PDEs) from data.
PDE discovery may be able to discover predictive models for physical systems that have proven difficult to model.

\bigskip

Crucially, any approach that aims to discover scientific laws must be able to work with scientific data.
Scientific data sets are often limited (sometimes only a few hundred data points) and may contain substantial noise. 
Therefore, PDE discovery algorithms must be able to work with limited, noisy data.
Further, scientists often perform multiple experiments on the same system.
As such, PDE discovery algorithms should be able to incorporate information from multiple data sets.
This paper presents a novel PDE discovery algorithm that meets these challenges.

\bigskip

{\bf Contributions:} In this paper, we develop a novel PDE discovery algorithm, \texttt{PDE-LEARN}, that can learn a general class of PDEs directly from data. 
In particular, \texttt{PDE-LEARN} can identify both linear and nonlinear PDEs with several spatial variables.
Significantly, \texttt{PDE-LEARN} is also robust to noise and limited data.
Further, \texttt{PDE-LEARN} can pool information from multiple experiments, by learning from multiple data sets simultaneously.
We confirm \texttt{PDE-LEARN}’s efficacy through a set of numerical experiments.
These experiments demonstrate that \texttt{PDE-LEARN} can discover a variety of linear and nonlinear PDEs, even when its training set is limited and noisy.

\bigskip 

{\bf Outline:} The rest of this paper is organized as follows: 
First, in section \ref{sec:Problem}, we state our assumptions on the data and the form of the hidden PDE.
Next, in section \ref{sec:Related_Work}, we survey related work on PDE discovery, focusing on methods that encode the hidden PDE into their loss functions.
In section \ref{sec:Methodology}, we give a comprehensive description of the \texttt{PDE-LEARN} algorithm.
In section \ref{sec:Experiments}, we demonstrate the efficacy of \texttt{PDE-LEARN} by discovering a variety of PDEs from limited, noisy data sets.
Section \ref{sec:Discussion} discusses our rationale behind the \texttt{PDE-LEARN} algorithm, practical considerations, limitations, and potential future research directions.
Finally, we provide concluding remarks in section \ref{sec:Conclusion}.

\section{Problem Statement} \label{sec:Problem}

Our algorithm uses noisy measurements of one or more system response functions to identify a \emph{hidden PDE} that those system response functions satisfy.
This section mainly serves to make the preceding statement more precise. 
First, we introduce the notation we use throughout this paper.
We then define what we mean by system response functions and state our assumptions about the hidden PDE they satisfy.
We then specify our assumptions about our noisy measurements of the system response function.
Finally, we conclude this section by formally stating the problem that our approach seeks to solve.

\bigskip

{\bf Problem domain:} Let $n_{D}, n_{S}\in \mathbb{N}$ and $\Omega_1, \ldots, \Omega_{n_{S}} \subseteq \mathbb{R}^{n_{D}}$ be a collection of open, connected sets. 
We refer to $\Omega_i$ as the \emph{$i$th spatial problem domain}.
We assume that for each $i \in \{1, 2, \ldots, n_S\}$, a physical process evolves on the domain $\Omega_i$ during the time interval $(0, T_i]$, for some $T_i > 0$.
Thus, there exists a \emph{system response function} $u_i : (0, T_i] \times \Omega_i \to \mathbb{R}$ that describes the state of the $i$th physical process at each position and time.
In particular, $u_i(t, X)$ denotes the system state at time $t \in (0, T_i]$ and position $X \in \Omega$.
We refer to the Cartesian product $(0, T_i] \times \Omega_i$ as the \emph{$i$th problem domain}.
If $n_S = 1$, we drop the subscript notation and refer to $(0, T] \times \Omega$ as \emph{the problem domain}.

\bigskip 

Roughly speaking, we assume that each system response function corresponds to an experiment.
Thus, by allowing multiple system response functions, we can deal with the case when the user has has data from several experiments.
It is also possible, and perfectly admissible, to have just one problem domain and response function.

\bigskip 

{\bf Derivative Notation:} In this paper, $D_{s}^{n} g$ denotes the $n$th partial derivative of any sufficiently smooth function $g$ with respect to the variable $s$.
Thus, for example, 

$$D_{x}^{2} g(t, x, y, z) = \partial^2 g(t, x, y, z) / \partial x^2.$$

For brevity, we abbreviate $D_{s}^1 g$ as $D_{s} g$. 
Throughout this paper, it will be helpful to have a concise notation for all partial derivative operators below a certain order.
With that in mind, let

$$\hat{\partial}^0 u, \hat{\partial}^1 u, \ldots, \hat{\partial}^{N_M} u$$

denote an enumeration of the partial derivatives of $u$ of order $\leq M$ (with the convention that the identity map is a $0$th-order partial derivative).
Note that $\hat{\partial}^0 u, \hat{\partial}^1 u, \ldots, \hat{\partial}^{N_M} u$ includes mixed partials if $u$ is a function of multiple variables.

\bigskip

As an example, let's consider the case when $M = 2$ and $n^D = 3$.
Thus, we are interested in all partial derivatives of order $\leq 2$ of a function of three variables, $x$, $y$, and $z$.
In this case, $N_M = 9$, and one possible enumeration is

$$\begin{aligned} \hat{\partial}^0 u = u,\qquad \hat{\partial}^1 u = D_x u,\qquad \hat{\partial}^2 u &= D_x^2 u,\qquad \hat{\partial}^3 u = D_y u, \qquad \hat{\partial}^4 u = D_y^2 u \\
\hat{\partial}^5 u = D_z u,\ \ \hat{\partial}^6 u = D_z^2 u,\qquad \hat{\partial}^7 u &= D_x D_y u,\ \  \hat{\partial}^8 u = D_y D_z u,\ \ \hat{\partial}^9 u = D_x D_z u. \end{aligned}$$

As this example demonstrates, $\hat{\partial}^5 u$ does not necessarily represent a fifth-order partial derivative of $u$. 
Rather, $\hat{\partial}^5 u$ denotes the fifth term in our \emph{specific enumeration} of the partial derivatives of $u$.

\bigskip 

{\bf Hidden PDE:} We assume there exists a \emph{hidden PDE} of order $\leq M$ such that $u_i$ satisfies the hidden PDE on $(0, T_i] \times \Omega_i$. 
We further assume this PDE takes the following form:

\begin{equation} f_0 \Big(\hat{\partial}^0 u_i, \ldots, \hat{\partial}^{N_M} u_i \Big) = \sum_{k = 1}^{K} c_k f_k\Big(\hat{\partial}^0 u_i, \ldots, \hat{\partial}^{N_M} u_i \Big). \label{eq:PDE} \end{equation}

Critically, we assume that equation \ref{eq:PDE} holds with \emph{the same coefficients} for each $i \in \{ 1, 2, \ldots, n_S \}$. 
In this expression, we refer to the functions $f_0, f_1, \ldots, f_K$ as the \emph{library terms}. 
We refer to $f_0$ as the \emph{left-hand side term} or \emph{LHS term} for short.
We also refer to $f_1, \ldots, f_K$ as the \emph{right-hand side terms} or \emph{RHS terms} for short.
Each library term is a function of $u$ and its partial derivatives (both in space and time) of order $\leq M$.

\bigskip 

At a high level, the algorithm we propose attempts to learn the coefficients $c_1, \ldots, c_K$ using noisy, limited measurements of the system response functions and an assumed set of library terms.
Before we can state our algorithm precisely, we need to make a few assumptions about the library and the system response data.

\bigskip 

{\bf Monomial Library Terms:} Many physical systems are governed by equation \ref{eq:PDE} for a particular $M$ and set of coefficients. 
In many cases of practical interest (\emph{e.g.} solid and fluid mechanics, thermodynamics, and quantum mechanics), the governing PDE consists of terms that are \emph{monomials} of $u$ and its partial derivatives.
That is, the terms are of the form

\begin{equation} f_k(\hat{\partial}^0 u, \ldots, \hat{\partial}^{N_M} u) = \prod_{m = 0}^{N_M} \left( \hat{\partial}^{m} u \right)^{p_k(m)}, \label{eq:Mon_Lib_Term} \end{equation}

For some $p_k(0), \ldots, p_{k}(N_M) \in \mathbb{N} \cup \{ 0 \}$.
We call library functions of this form \emph{monomial library functions}.
In this paper, we consider libraries consisting of monomial library terms.
In principle, however, our proposed algorithm can work with any library.

\bigskip 

{\bf Data Points:} Let $i \in \{ 1, 2, \ldots, n_S \}$ and let $\left\{ \left(t_{j}^{(i)}, X_{j}^{(i)} \right) \right\}_{j = 1}^{N_{Data}(i)} \subseteq (0, T] \times \Omega$ be a collection of \emph{data points} in the $i$th problem domain.
We assume that we have noisy measurements of $u_i$ at these data points, which we denote by

$$\left\{ \Tilde{u_i}\left(t_j^{(i)}, X_j^{(i)}\right) \right\}_{j = 1}^{N_{Data}(i)} \subseteq (0, T_i] \times \Omega_i.$$ 

We refer to the collection of these measurements as the \emph{$i$th noisy data set}.
Further, we refer to the corresponding set $\left\{ u_i\left(t_j^{(i)}, X_j^{(i)}\right) \right\}_{j = 1}^{N_{Data}(i)}$ as the \emph{$i$th noise-free data set}.
Critically, we only assume knowledge of the noisy data set.
In general, since the measurements are noisy, 

$$u_i\left(t_j^{(i)}, X_j^{(i)}\right) \neq \tilde{u}_i \left(t_j^{(i)}, X_j^{(i)} \right).$$

We assume, however, that for each $j \in \{1, 2, \ldots, N_{Data}(i) \}$, 

$$u\left( t_j^{(i)}, X_j^{(i)} \right) - \tilde{u}_i \left(t_j^{(i)}, X_j^{(i)} \right) \sim N(0, \sigma_i^2),$$

for some $\sigma_i > 0$.
We refer to this difference as the \emph{noise} at the data point $\left(t_j^{(i)}, X_j^{(i)} \right)$.
We assume that the noises at different data points are independent and identically distributed.
We define the \emph{noise level} of a data set as the ratio of $\sigma_i$ to the standard deviation of the noise-free data set.
Finally, if $n_S = 1$, we write $N_{Data}$ for $N_{Data}(1)$.

\bigskip

{\bf Goals:} Our goal is to use the noisy data sets, $\left\{ \Tilde{u_i}\left(t_j^{(i)}, X_j^{(i)} \right) \right\}_{j = 1}^{N_{Data}(i)}$, and the library, $f_0, \ldots, f_K$, to learn the coefficients $c_1, \ldots, c_K$ in equation \ref{eq:PDE}.
To do this, we learn an approximation to each $u_i$ that satisfies equation \ref{eq:PDE} for some set of coefficients.
To maximize the applicability of our approach, we make as few assumptions as possible about the hidden PDE.
In particular, we generally use a large library.
The goal here is to select a library that is broad enough to include the terms that are present (have non-zero coefficients) in the hidden PDE without requiring the user to identify those terms beforehand.
This approach means our library includes many extraneous terms; i.e., most of the coefficients should be zero.
Therefore, we tacitly assume the right-hand side of equation \ref{eq:PDE} is sparse.

\bigskip

For reference, table \ref{Table:Problem:Notation} lists the notation we introduced in this section.

\bigskip

\begin{table}[hbt]
    \centering 
    \rowcolors{2}{white}{cyan!10}
    
    \begin{tabulary}{1.0\linewidth}{p{3.1cm}L}
        \toprule[0.3ex]
        \textbf{Notation} & \textbf{Meaning} \\
        \midrule[0.1ex]
        $n_{D}$ & The number of spatial dimensions in the spatial problem domain. \\
        \addlinespace[0.4em]
        $n_{S}$ & The number of system response functions. \\
        \addlinespace[0.4em]
        $\Omega_i$ & The $i$th spatial domain\footnotemark: an open, connected subset of $\mathbb{R}^{n_{D}}$. \\
        \addlinespace[0.4em]
        $(0, T_i] \subseteq \mathbb{R}$ & The time interval over which $u_i$ evolves on $\Omega_i$. \\
        \addlinespace[0.4em]
        $u_i : (0, T_i] \times \Omega_i \to \mathbb{R}$ & The $i$th system response function. If $n_S = 1$, we denote $u_1 = u$. \\
        \addlinespace[0.4em]
        $N_M$ & The number of distinct partial derivatives of $u$ of order $\leq M$. \\
        \addlinespace[0.4em]
        $\hat{\partial}^0, \ldots, \hat{\partial}^{N_M}$ & An enumeration of the partial derivatives of order $\leq M$ (including the identity map). \\
        \addlinespace[0.4em]
        $D_{s}^{n} g$ & the $n$th derivative of some function $g$ with respect to the variable $s$. \\
        \addlinespace[0.4em]
        $D_{s} g$ & abbreviated notation for $D_{s}^1 g$. \\
        \addlinespace[0.4em]
        $N_{Data}(i)$ & Number of data points in the $i$th noisy-data set. If $n_S = 1$, we write $N_{Data}(1) = N_{Data}$. \\  
        \addlinespace[0.4em]
        $\left\{ \left( t_j^{(i)}, X_j^{(i)} \right) \right\}_{j = 1}^{N_{Data}\left(i\right)}$ & The data points for the $i$th system response function. \\
        \addlinespace[0.4em]
        $\tilde{u}_i(t, X)$ & A noisy measurement of $u_i$ at $(t, X) \in (0, T_i] \times \Omega_i$. \\
        \addlinespace[0.4em]
        $K$ & The number of library functions. See equation \ref{eq:PDE} \\
        \addlinespace[0.4em]
        $f_0$ & The left-hand side term. See equation \ref{eq:PDE}. \\
        \addlinespace[0.4em]
        $f_1 \ldots, f_K$ & The right-hand side terms. See equation \ref{eq:PDE}. \\
        \addlinespace[0.4em]
        $c_1, \ldots, c_K$ & The coefficients of the RHS terms terms $f_1, \ldots, f_K$ in equation \ref{eq:PDE}. \\
        \addlinespace[0.4em]
        Noise level & The ratio of the standard deviation of the noise to that of the noise-free data set. \\
        \bottomrule[0.3ex]
    \end{tabulary}
    
    \caption{The notation and terminology of section (\ref{sec:Problem})} 
    \label{Table:Problem:Notation}
\end{table}

\footnotetext{For a given problem, all spatial domains come from the SAME Euclidean subspace. That is, $\Omega_i \subseteq \mathbb{R}^{n_{D}}$ for each $i$.}

\section{Related Work}
\label{sec:Related_Work}

In this section, we discuss relevant previous work on PDE discovery to contextualize our contributions.
PDE discovery began in the late 2000s with two papers, \cite{bongard2007automated} and \cite{schmidt2009distilling}.
While neither paper is directly concerned with identifying PDEs (the former focuses on discovering dynamical systems, while the latter focuses on identifying invariant and conservation laws), they set the stage for using machine learning to discover scientific laws from data.
Both approaches use genetic algorithms to learn relationships (hidden dynamical system in the former and conservation laws in the latter) that the system response function satisfies.

\bigskip 

A significant advance came a few years after \cite{schmidt2009distilling} when Rudy et al. introduced \texttt{PDE-FIND} \cite{rudy2017data}. 
Developed as a modification of the Sparse Identification of Nonlinear DYnamics (SINDY) algorithm \cite{brunton2016discovering}, \texttt{PDE-FIND} represents one of the earliest and most important breakthroughs in identifying PDEs directly from data. 
\texttt{PDE-FIND} uses similar assumptions to the ones listed in section \ref{sec:Problem} but additionally assumes that $n_{S} = 1$ (a single experiment), $f_0(u) = D_t u$, and the RHS terms depend only on the lone system response function, $u$, and its spatial partial derivatives.
Critically, \texttt{PDE-FIND} also assumes the data points occur on a regular grid.
This additional constraint allows \texttt{PDE-FIND} to use numerical differentiation techniques to approximate the partial derivatives of $u$.
Using these approximations, \texttt{PDE-FIND} can evaluate the library terms at the data points, which engenders a linear system for the coefficients $c_1, \ldots, c_K$.
\texttt{PDE-FIND} then finds a sparse, approximate solution to this system using an algorithm called \emph{Sequentially Thresholded Least Squares}, or \texttt{ST-Ridge} for short.
\texttt{PDE-FIND} can successfully identify a wide range of PDEs directly from data.
With that said, \texttt{PDE-FIND} does have some fundamental limitations.
In particular, since numerical differentiation tends to amplify noise, \texttt{PDE-FIND}'s performance decreases considerably in the presence of moderate noise levels.
Further, requiring the data to occur on a regular grid is a cumbersome limitation for scientific applications, where data can be difficult and expensive to acquire. 

\bigskip

In addition to \texttt{PDE-FIND}, two other notable examples of early PDE discovery algorithms are \cite{schaeffer2017learning} and \cite{berg2017neural}.
The former is similar to \texttt{PDE-FIND} but uses spectral methods to approximate the derivatives.
This change enables their approach to identify a variety of PDEs, even in the presence of significant noise.
Like \texttt{PDE-FIND}, however, \cite{schaeffer2017learning} does require that the data points occur on a regular grid.
The latter trains a neural network, $U : (0, T] \times \Omega \to \mathbb{R}$, to match a noisy data set (thereby learning an approximation to the system response function) and then uses sparse regression to identify the coefficients $c_1, ... , c_K$.
Using a network to interpolate the data allows the data points to be dispersed anywhere in the problem domain.

\bigskip

Another significant contribution to PDE discovery came a few years later with \texttt{DeepMoD} \cite{both2021deepmod}.
\texttt{DeepMoD} learns an approximation, $\xi$, to the coefficients $c_1, ... , c_K$, while simultaneously training a neural network, $U : (0, T] \times \Omega \to \mathbb{R}$, to match a noisy data set.
Thus, their approach learns the hidden PDE and the system response function at the same time.
Further, like \cite{berg2017neural}, the data points for \texttt{DeepMoD} can be arbitrarily distributed throughout the problem domain.
\texttt{DeepMoD} uses \emph{Automatic Differentiation} \cite{baydin2018automatic} to calculate the partial derivatives of the neural network at randomly selected \emph{collocation points} in the problem domain.
\texttt{DeepMoD} can then use these partial derivatives to evaluate the library terms at the collocation points. 
To train $U$ and $\xi$, \texttt{DeepMoD} uses a three-part loss function.
The first part measures how well $U$ matches the data set, the second measures how well $U$ satisfies the hidden PDE \ref{eq:PDE} with the components of $\xi$ in place of $c_1, \ldots, c_K$, and the third is the $L^1$ norm of $\xi$ (this promotes sparsity in $\xi$). 
The second and third parts of the loss function embed the LASSO loss function within \texttt{DeepMoD}'s loss function.
These parts encourage $U$ to learn the function that roughly matches the data but also satisfies a PDE of the form of equation \ref{eq:PDE}.
This approach embeds the fact that the system response function satisfies a PDE characterized by $\xi$ into the loss function.
This loss function produces a robust algorithm that can identify PDEs, even from noisy and limited data. 
\texttt{DeepMoD} served as the original inspiration for \texttt{PDE-LEARN}.
Finally, it is worth noting that \cite{chen2021physics} proposed an approach similar to \texttt{DeepMoD} but uses an innovative training scheme that achieves impressive results on several PDEs.

\bigskip

More recently, the authors of this paper proposed another algorithm, \texttt{PDE-READ} \cite{stephany2022pde}. 
\texttt{PDE-READ} uses two neural networks: The first, $U : (0, T] \times \Omega \to \mathbb{R}$, learns the system response function, and the second, $N$, learns an abstract representation of the right-hand side of equation \ref{eq:PDE}. 
That approach is based on Raissi’s \emph{deep hidden physics models} algorithm \cite{raissi2018deep}. 
Like \texttt{DeepMoD}, \texttt{PDE-READ} learns an approximation of the system response function while simultaneously identifying the hidden PDE.
Significantly, \texttt{PDE-READ} utilizes \emph{Rational Neural Networks} \cite{boulle2020rational}, a type of fully connected neural network whose activation functions are trainable rational functions. 
After training both networks, \texttt{PDE-READ} uses a modified version of the \emph{Recursive Feature Elimination} algorithm \cite{guyon2002gene} to extract $c_1, \ldots, c_K$ from $N$.
This approach proves impressively robust, as \texttt{PDE-READ} can identify a variety of PDEs even from limited measurements with very high noise levels.

\bigskip

The methods discussed above mostly use a standard fully connected neural network to approximate the system response function and identify the hidden PDE. 
The PDE-discovery community has, however, proposed many other approaches.
\cite{gurevich2019robust} and \cite{messenger2021weak} learn the hidden PDE via a weak-formulation approach.
Using weak forms places additional restrictions on the form of the hidden PDE but engenders an algorithm that is remarkably robust to noise. 
Further, \cite{atkinson2019data} uses a Gaussian process to approximate the system response function and a genetic algorithm to identify the hidden PDE. 
Finally, \cite{bonneville2021bayesian} uses Bayesian Neural networks to learn the system response function and identify the hidden PDE.

\bigskip 

In this paper, we introduce a new PDE discovery algorithm, \texttt{PDE-LEARN}.
\texttt{PDE-LEARN} can learn a broad class of PDEs directly from noisy and limited data.
Like \texttt{PDE-READ}, \texttt{PDE-LEARN} utilizes Rational Neural Networks \cite{boulle2020rational} to learn an approximation of the system response functions.
Further, like \texttt{DeepMoD}, \texttt{PDE-LEARN} uses a sparse, trainable vector, $\xi$, to learn an approximation to the coefficients $c_1, \ldots, c_K$. 
What sets \texttt{PDE-LEARN} apart, however, is its three-part loss function that incorporates aspects of the \textit{iteratively reweighted least-squares} algorithm \cite{chartrand2008iteratively} to help promote sparsity in $\xi$. 
Equally important, \texttt{PDE-LEARN} uses an adaptive procedure to place additional collocation points in regions where the collocation loss is the greatest. 
This process helps accelerate convergence and yields an algorithm that is highly effective in the low-data, high-noise data limit.

\section{Methodology} \label{sec:Methodology}

In this section, we describe our algorithm - {\it PDE discovery via $L^0$ Error Approximation and Rational Neural networks} - or \texttt{PDE-LEARN} for short.
\texttt{PDE-LEARN} does the following: 
First, it uses noisy data sets, $\{ \Tilde{u_i}(t_j^{(i)}, X_j^{(i)}) \}_{j = 1}^{N_{Data}(i)}$, to learn an approximation to each system response function, $u_i$.
Second, it uses the fact that the system response functions satisfy a PDE of the form of equation \ref{eq:PDE} to learn the coefficients $c_1, \ldots, c_{K}$.
Significantly, \texttt{PDE-LEARN} can learn any PDE of the form of equation \ref{eq:PDE} and can operate with multiple spatial variables.

\bigskip

\texttt{PDE-LEARN} uses a Rational Neural Network \cite{boulle2020rational}, $U_i : (0, T] \times \Omega \to \mathbb{R}$, to approximate the $i$th system response function, $u_i$.
It approximates the coefficients $c_1, \ldots, c_K$ in equation \ref{eq:PDE} using a trainable vector, $\xi$.
During training, \texttt{PDE-LEARN} learns $\xi$ and each $U_i$ by minimizing the loss function

\begin{equation} \text{Loss}\left( U_1, \ldots, U_{n_S}, \xi \right) = w_{Data}\sum_{i = 1}^{n_S}\text{Loss}_{\text{Data}}(U_i) + w_{Coll}\sum_{i = 1}^{n_S} \text{Loss}_{\text{Coll}}(U_i, \xi) + w_{L^p}\text{Loss}_{L^p}(\xi). \label{eq:Loss} \end{equation}

Here, $w_{Data}$, $w_{Coll}$, and $w_{L^p}$ are user-selected scalar hyperparameters.
Further,

\begin{align} &\text{Loss}_{\text{Data}}(U_i) = \left( \frac{1}{N_{Data}(i)} \right) \sum_{j = 1}^{N_{Data}(i)} \left| U\left(t_j^{(i)}, X_j^{(i)} \right) - \tilde{u}\left(t_j^{(i)}, X_j^{(i)} \right) \right|^2 \label{eq:Loss:Data} \\
&\text{Loss}_{\text{Coll}}(U_i, \xi) = \left( \frac{1}{N_{Coll}(i)} \right) \sum_{j = 1}^{N_{Coll}(i)} \left| R_{PDE}\left(U_i, \hat{t}_j^{(i)}, \hat{X}_j^{(i)}\right) \right|^2 \label{eq:Loss:Coll} \\
&\text{Loss}_{L^p}(\xi) = \sum_{k = 1}^{K} a_k \xi_k^2, \label{eq:Loss:Lp} \end{align}

In equation \ref{eq:Loss:Lp}, \texttt{PDE-LEARN} updates the constants $a_k$ at the start of each epoch such that the $L^p$ loss approximates the $p$ norm of $\xi$. 
We discuss this in detail in section \ref{sub_sec:Method:Lp}.
In equation \ref{eq:Loss:Coll}, $R_{PDE}$ is the \emph{PDE-Residual}, defined by

\begin{equation}R_{PDE}\left(U_i, \hat{t}_j^{(i)}, \hat{X}_j^{(i)} \right) = f_0 \left(\hat{t}_i, \hat{X}_i \right) - \sum_{k = 1}^{K} \xi_k f_k\left(\hat{t}_i, \hat{X}_i\right), \label{eq:PDE:Residual}\end{equation}

where $f_k\left(\hat{t}_j^{(i)}, \hat{X}_j^{(i)}\right)$ is an abbreviation for $f_k\left(\hat{\partial}^0 U\left(\hat{t}_j^{(i)}, \hat{X}_j^{(i)} \right), \ldots, \hat{\partial}^{N_M} U\left(\hat{t}_j^{(i)}, \hat{X}_j^{(i)} \right) \right)$. 
\texttt{PDE-LEARN} uses \emph{automatic differentiation} \cite{baydin2018automatic} to calculate the partial derivatives of $U$ and subsequently evaluate the library functions.
The points $\left\{ \left( \hat{t}_i^{(j)}, \hat{X}_i^{(j)} \right) \right\}_{i = 1}^{N_{Coll}(i)} \subseteq (0, T_i] \times \Omega_i$ are the \emph{collocation points}.
We discuss these in detail below in section \ref{sub_sec:Method:Collocation}.

\bigskip

\begin{figure}
    \centering
    \includegraphics[width=\linewidth]{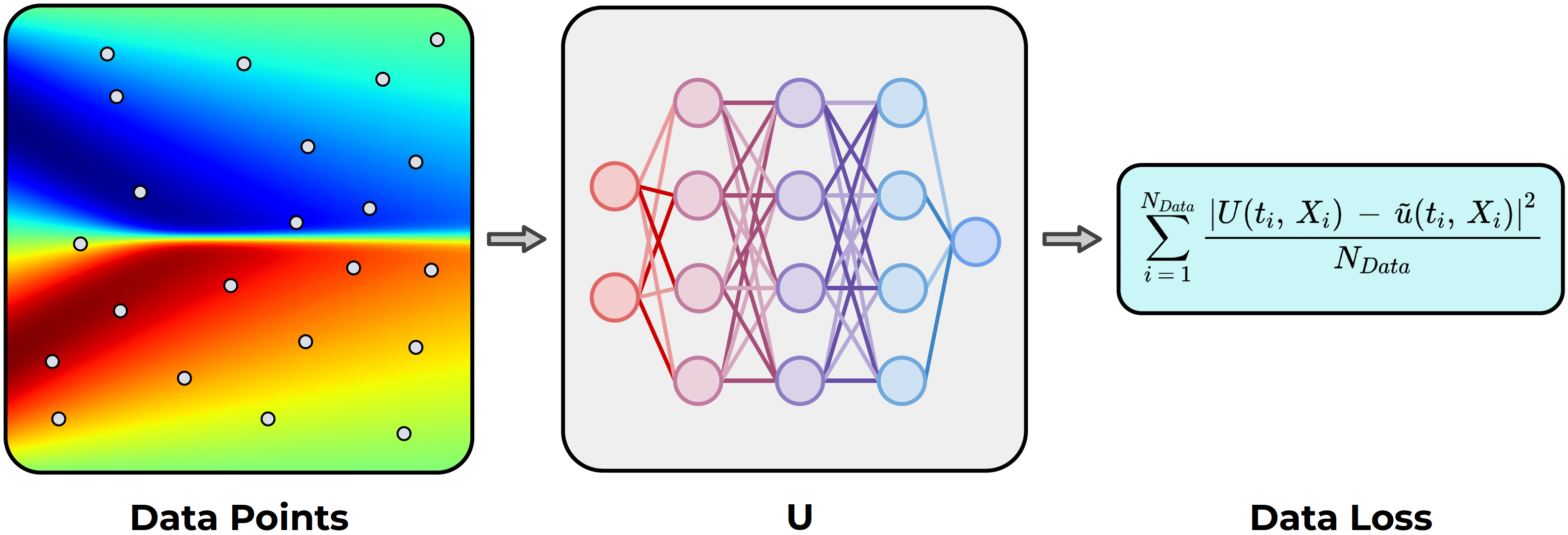}
    \caption{This figure depicts the process that \texttt{PDE-LEARN} uses to evaluate the data loss. 
    The white circles on the left side of the figure represent the data points.
    Moving from left to right, we feed data points to the rational neural network, $U$. 
    We then compare the resulting predictions with the noisy measurements of the system response function (in this case, for Burger's equation in section \ref{sub_sec:Burgers}).
    This process yields the data loss (right side of the figure).}
    \label{fig:Loss:Data}
\end{figure}

We refer to $Loss_{Data}$, $Loss_{Coll}$, and $Loss_{L^p}$ as the \emph{Data}, \emph{Collocation}, and \emph{L\textsuperscript{p}} losses, respectively. 
Figures \ref{fig:Loss:Data} and \ref{fig:Loss:Coll} depict how \texttt{PDE-LEARN} evaluates the Data and Collocation Losses, respectively. 

\bigskip

The Data Loss forces $U$ to satisfy the noisy data, $\{ \Tilde{u}(t_i, X_i) \}_{i = 1}^{N_{Data}}$ at the data points. 
It is the mean square error between $U$'s predictions and the noisy data set.
The collocation loss forces $U$ to satisfy the PDE encoded in

$$ f_0 \left(\hat{t}_j^{(i)}, \hat{X}_j^{(i)} \right) = \sum_{k = 1}^{K} \xi_k f_k\left(\hat{t}_j^{(i)}, \hat{X}_j^{(i)} \right),$$

at the collocation points $\left\{ \left( \hat{t}_j^{(i)}, \hat{X}_j^{(i)} \right) \right\}_{i = 1}^{N_{Coll}(i)}$. 
It is what couples the training of $U$ and $\xi$.
Finally, $\text{Loss}_{L^p}$ encodes our assumption that most of the coefficients in equation \ref{eq:PDE} are zero by promoting sparsity in $\xi$.
It is a weighted sum of the squares of components of $\xi$.
\texttt{PDE-LEARN} accomplishes this by re-selecting the weights, $a_k$, at the start of each epoch.
We discuss this in detail in section \ref{sub_sec:Method:Lp}.
\bigskip

To use \texttt{PDE-LEARN}, one must provide a noisy data set, select an architecture for $U$, and select an \emph{appropriate} collection of library terms.
Here, \emph{appropriate} means that the right-hand side of the hidden PDE can be expressed as a sparse linear combination of the library terms.
Once \texttt{PDE-LEARN} has finished training, it reports identified PDE.

\subsection{Collocation Loss}
\label{sub_sec:Method:Collocation}
In equation \ref{eq:PDE:Residual}, $R_{PDE}\left(U_i, \hat{t}_j^{(i)}, \hat{X}_j^{(i)} \right)$ is the \emph{PDE residual} of $U_i$ at the \emph{collocation point} $\left(\hat{t}_j^{(i)}, \hat{X}_j^{(i)}\right) \in (0, T] \times \Omega$.
\texttt{PDE-LEARN} uses two types of collocation points: \emph{random collocation points} and \emph{targeted collocation points}.
Each problem domain has collocation points (both random and targeted).
For each problem domain, \texttt{PDE-LEARN} selects the random collocation points by repeatedly sampling from a uniform distribution over the problem domain,

$$\left( \hat{t}_j^{(i)}, \hat{X}_j^{(i)} \right) \sim \text{Unif}\big((0, T_i] \times \Omega_i \big).$$

The number of random collocation points in each problem domain, $N_{Coll\ Random}$, is a hyperparameter.
\texttt{PDE-LEARN} re-samples the random collocation points for each problem domain at the start of each epoch.
The targeted collocation points are the random collocation points from previous epochs at which the PDE residual is unusually large. 
At the start of training, we initialize the targeted collocation points to be the empty set.
During subsequent epochs, \texttt{PDE-LEARN} uses the following procedure for each system response function:

\begin{enumerate}
\item During each epoch \texttt{PDE-LEARN} records the absolute value of the PDE residual at each collocation point (both random and targeted) for the $i$th system response function.
PDE-LEARN records this set of non-negative values.
\item \texttt{PDE-LEARN} then computes the mean and standard deviation of this set.
\item \texttt{PDE-LEARN} then determines which collocation points have an absolute PDE-Residual that is more than three standard deviations larger than the mean. 
These points are the targeted collocation points for the next epoch.

\end{enumerate}

This procedure accelerates training by adaptively focusing the training of $U_i$ and $\xi$ on regions of the problem domain where the PDE residual is large.
If the PDE residual is large in a particular region of the $i$th problem domain, any collocation points in that region will become targeted collocation points.
Because the random collocation points are re-sampled, new collocation points will appear in the problematic region at the start of each epoch.
Thus, targeted collocation points will accumulate in that region.
Eventually, the PDE residual in that region will dominate the collocation loss.
This forces $U_i$ and $\xi$ to adjust until the PDE residual in the problematic region shrinks.

\bigskip 

\begin{figure}
    \centering
    \makebox[\textwidth][c]{\includegraphics[width=1.15\textwidth]{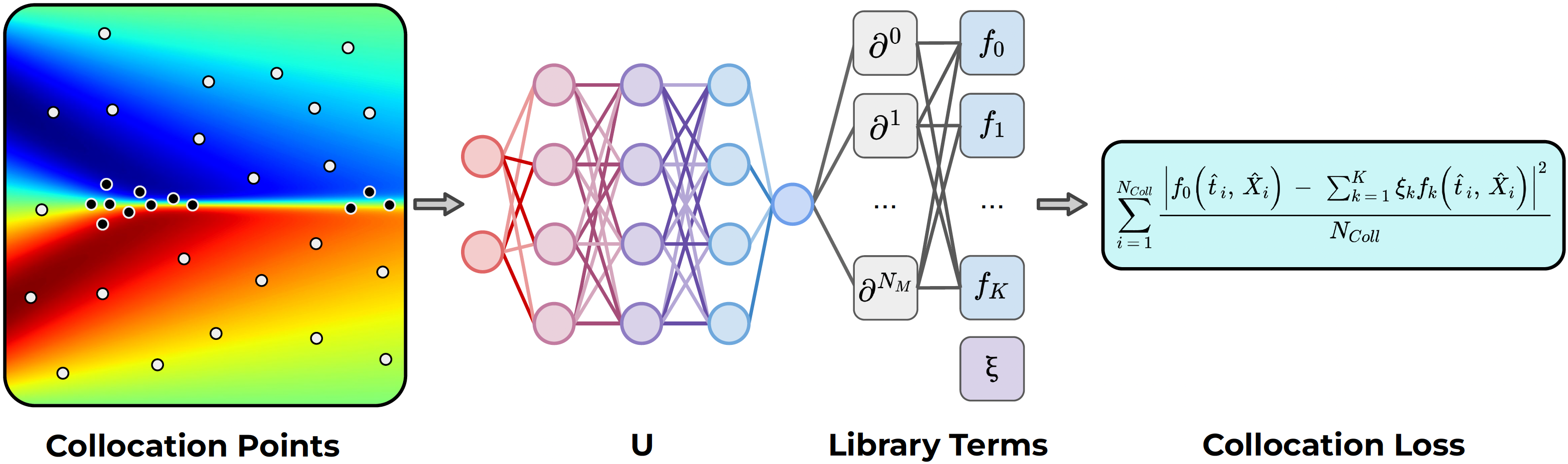}}
    \caption{This figure depicts the process that \texttt{PDE-LEARN} uses to calculate the collocation loss.
    The white and black circles on the left side of the figure represent the random and targeted collocation points, respectively.
    Moving from left to right, \texttt{PDE-LEARN} evaluates $U$ at each collocation point.
    It then uses automatic differentiation to evaluate $\hat{\partial}^0 U$, \ldots, $\hat{\partial}^N_M U$, and the library functions at the collocation points.
    Finally, using these values and $\xi$, \texttt{PDE-LEARN} evaluates the Collocation Loss.}
    \label{fig:Loss:Coll}
\end{figure}

Though \texttt{PDE-LEARN} can work with an arbitrary collection of library functions, we only consider monomial library functions in this paper.
To evaluate these functions efficiently, \texttt{PDE-LEARN} records every partial derivative operator that is present in at least one library function (equivalently, $\{ \hat{\partial}_j : \exists\ k \in \{ 0, 1, 2, \ldots, K \} \text{ such that } \hat{\partial}_j u \text{ is one of the sub-terms of } f_k \}$).
At the start of each epoch, \texttt{PDE-LEARN} evaluates these partial derivatives of each $U_i$ at each collocation point. 
We implemented this process to be as efficient as possible.
In particular, it starts by computing the lowest-order partial derivatives of $U_i$.
For subsequent partial derivatives, \texttt{PDE-LEARN} uses the following rule: if we can express a partial derivative of $U_i$ as a partial derivative operator applied to another partial derivative of $U_i$ that we have already computed, then compute the new partial derivative from the old one.
This approach allows us to compute all the necessary partial derivatives of $U_i$ without any redundant computations. 
After computing the partial derivatives of $U_i$, \texttt{PDE-LEARN} uses them to evaluate the library terms at the collocation points. 
It then evaluates the PDE-Residual, equation \ref{eq:PDE:Residual}, at each collocation point.
Finally, from the PDE residuals, \texttt{PDE-LEARN} can compute the collocation loss, equation \ref{eq:Loss:Coll}.
Figure \ref{fig:Loss:Coll} depicts the process that \texttt{PDE-LEARN} uses to calculate the Collocation Loss.

\bigskip

\subsection{L\textsuperscript{p} Loss} 
\label{sub_sec:Method:Lp}

The collocation and $L^p$ losses embed the \emph{iteratively reweighted least-squares} \cite{chartrand2008iteratively} loss function into ours. 
At the start of each epoch, \texttt{PDE-LEARN} updates the weights $a_1, \ldots, a_K$ in equation \ref{eq:Loss:Lp} using the following rule:

\begin{equation} a_k = \left( \frac{1}{\min\{\delta, |\xi_k|^{2 - p} \} } \right) \label{eq:Weights} \end{equation} 

Here, $\delta > 0$ is a small constant to prevent division by zero, and $p \in (0, 2)$ is a hyperparameter.
Crucially, the value $\xi_k$ that appears in this equation is the $k$th component of $\xi$ at the \emph{start} of the epoch; thus $a_k$ is treated as a constant during backpropagation.
Notably, if $\xi_k$ is sufficiently large, then

$$ a_k \xi_k^2 = |\xi_k|^p$$

and so, 

$$ \text{Loss}_{L^p} = \sum_{k = 0}^{K} |\xi_k|^p \approx \| \xi \|_{0}.$$

Critically, \texttt{PDE-LEARN} evaluates each $a_k$ at the start of every epoch and treats them as constants during that epoch.
This subtle detail allows $\text{Loss}_{L^p}$ to closely approximate $\| \xi \|_{0}$ while remaining a smooth, convex function of $\xi$'s components. 
We discuss this in section \ref{sub_sec:Discussion:Lp_Loss}.

\bigskip

\subsection{Training}
\label{sub_sec:Training}
\texttt{PDE-LEARN} identifies the hidden PDE using a three-step training process.
We use the Adam \cite{kingma2014adam} optimizer to minimize \ref{eq:Loss} in all three steps.

\bigskip

We call the first step the \emph{burn-in} step.
\texttt{PDE-LEARN} first initializes $\xi$ and the $U_i$'s. 
It initializes $\xi$ to a vector of zeros.
It initializes the weights matrices and bias vectors in $U_i$ using the initialization procedure in \cite{glorot2010understanding}.
Finally, it initializes the rational activation functions using the procedure in \cite{boulle2020rational}.
\texttt{PDE-LEARN} then sets $w_{L^p}$ to zero and begins training. 
During this step, $U_i$ learns an approximation to $u_i$.
Since $w_{L^p} = 0$, almost all components of $\xi$ become non-zero; we do not attempt to identify the PDE during this step.

\bigskip

At the end of the burn-in step, \texttt{PDE-LEARN} \emph{prunes} $\xi$ by eliminating all RHS terms whose corresponding component of $\xi$ is smaller than a threshold.
Throughout this paper, we select the threshold to be slightly larger than $\sqrt{\varepsilon}$, where $\varepsilon$ is machine epsilon for single-precision floating numbers.
We discuss the implications of pruning in section \ref{sub_sec:Discussion:Pruning}.

\bigskip

In the second step, which we call the \emph{sparsification} step, we set $w_{L^p}$ to a small, non-zero value and then resume training.
During this step, $\xi$ becomes sparse, only retaining the components of $\xi$ that correspond to RHS terms that are present in the hidden PDE.
After training, we repeat the pruning process (which usually eliminates the bulk of the extraneous RHS terms).
By the end of this step, \texttt{PDE-LEARN} identifies which RHS terms have non-zero coefficients.
However, since the $L^p$ loss encourages each coefficient to go to zero, the magnitudes of the retained coefficients are usually too small at this point.

\bigskip

In the third step, which we call the \emph{fine-tuning} step, we set $w_{L^p}$ to zero once again and resume training.
This step retains only the RHS terms that survived the sparsification step.
By removing the $L^p$ loss, the components of $\xi$ are no longer under pressure to be as close to $0$ as possible, which allows them to converge to the values in the hidden PDE.
\texttt{PDE-LEARN} trains until the $L^p$ loss stops increasing.
\texttt{PDE-LEARN} then reports the identified PDE encoded in $\xi$. 

\bigskip

For brevity, we will let $N_{Burn-in}$, $N_{Sparse}$, and $N_{Fine-tune}$ denote the number of burn-in, sparsification, and fine-tuning epochs, respectively

\bigskip

Table \ref{Table:Method:Notation} lists the notation we introduced in this section.

\begin{table}[hbt]
    \centering 
    \rowcolors{2}{white}{cyan!10}
    
    \begin{tabulary}{\linewidth}{p{3cm}L}
        \toprule[0.3ex]
        \textbf{Notation} & \textbf{Meaning} \\
        \midrule[0.1ex]
        $U_i$ & Neural Network to approximate $u_i$. \\
        \addlinespace[0.4em]
        $\xi$ & A trainable vector in $\mathbb{R}^K$ whose components approximate $c_1, \ldots, c_K$. See equation \ref{eq:PDE}.  \\
        \addlinespace[0.4em]
        $p$ & A hyperparameter representing the ``p'' in ``$L^p$''. See equation \ref{eq:Loss:Lp}. \\
        \addlinespace[0.4em]
        $Loss_{Data}$ & The data loss. It measures the mean square error between $U_i$ and the noisy measurements of $u_i$ at the $i$th data points. See equation \ref{eq:Loss:Data} \\
        \addlinespace[0.4em]
        $Loss_{Coll}$ & The collocation loss. It measures how well $U$ satisfies the hidden PDE encoded in $\xi$ at the collocation points. See equation \ref{eq:Loss:Coll}. \\
        \addlinespace[0.4em]
        $Loss_{L^p}$ & The $L^p$ loss. It represents the $L^p$ quasi-norm of $\xi$ raised to the $p$th power. See equation \ref{eq:Loss:Lp}. \\
        \addlinespace[0.4em]
        $R_{PDE}$ & The PDE-residual. See equation \ref{eq:PDE:Residual}. \\
        \addlinespace[0.4em]
        $f_k\left(\hat{t}_j^{(i)}, \hat{X}_j^{(i)} \right)$ & An abbreviation of $f_k\left(\hat{\partial}^0 U\left(\hat{t}_j^{(i)}, \hat{X}_j^{(i)} \right), \ldots, \hat{\partial}^{N_M} U\left(\hat{t}_j^{(i)}, \hat{X}_j^{(i)} \right) \right)$. \\
        \addlinespace[0.4em]
        $w_{Data}, w_{Coll}, w_{L^p}$ & Hyperparameters that specify the weight of $Loss_{Data}$, $Loss_{Coll}$, and $Loss_{L^p}$, respectively. See equation \ref{eq:Loss}. \\
        \addlinespace[0.4em]
        $N_{Coll}(i)$ & The number of collocation points for $u_i$. See equation \ref{eq:Loss:Coll}. \\
        \addlinespace[0.4em]
        $N_{Random\ Coll}$ & A hyperparameter specifying the number of random collocation points. We use the same value for each system response function. See sub-section \ref{sub_sec:Method:Collocation}. \\
        \addlinespace[0.4em]
        $\left\{ \left(\hat{t}_j^{(i)}, \hat{X}_j^{(i)} \right) \right\}_{i = 1}^{N_{Coll}(i)}$ & The $i$th set of collocation points. See equation \ref{eq:Loss:Coll}. \\
        \addlinespace[0.4em]
        $N_{Burn-in}$ & The number of burn-in epochs. \\
        \addlinespace[0.4em]
        $N_{Sparse}$ & The number of sparsification epochs. \\
        \addlinespace[0.4em] 
        $N_{Fine-tune}$ & The number of fine-tuning epochs. \\
        \bottomrule[0.3ex]
    \end{tabulary}
    
    \caption{The notation and terminology of section (\ref{sec:Methodology})} 
    \label{Table:Method:Notation}
\end{table}

\bigskip

\section{Experiments} 
\label{sec:Experiments}

We implemented \texttt{PDE-LEARN} as an open-source Python library. 
Our implementation is publicly available
at \url{https://github.com/punkduckable/PDE-LEARN}, along with auxiliary MATLAB scripts to generate our data sets.

\bigskip 

As stated in section \ref{sec:Methodology}, each $U_i$ is a rational neural network \cite{boulle2020rational}. 
Thus, $U_i$'s activation functions are trainable $(3, 2)$ rational functions ($3$rd order polynomial in the numerator, second-order polynomial in the denominator). 
In other words, the coefficients that define the numerator and denominator polynomials are trainable parameters that the network learns along with its weight matrices and bias vectors.
Each hidden layer gets its own activation function, which we apply to each hidden unit in that layer.

\bigskip

In this section, we test \texttt{PDE-LEARN} on several PDEs, both linear and non-linear.
All of the data we use in these experiments come from numerical simulations.
The data sets from these simulations represent our noise-free data sets.
To create noisy, limited data sets with $N_{Data} \in \mathbb{N}$ points and a noise level $q \geq 0$, we use the following procedure:
\begin{enumerate}
    \item Calculate the standard deviation, $\sigma_{nf}$, the samples of the system response function in the noise-free data set.
    \item Select a subset of size $N_{Data}$ from the noise-free data set by sampling $N_{Data}$ points from the noise-free data set without replacement. The resulting subset is the \emph{limited data set}.
    \item Independently sample a Gaussian Distribution with mean $0$ and standard deviation $q*\sigma_{nf}$ once for each point in the limited data set. Add the $i$th value to the $i$th point in the limited data set. The resulting set is our noisy, limited data set.
\end{enumerate}

\bigskip

As discussed in section \ref{sub_sec:Training}, we use a three-step training process to train $\xi$ and $\{ U_1, \ldots, U_{n_S} \}$.
In our experiments, we stop the burn-in step when loss stops decreasing, which often takes between $1000$ and $1500$ epochs.
For the sparsification step, we select a small $w_{L^p}$ value (usually $0.0001$) and train until the $L^p$ loss stabilizes for a few hundred epochs (usually $1,000$ to $2,000$ epochs after burn-in).
Finally, for the fine-tuning step, we stop training once the $L^p$ loss stops increasing or once an equation-specific\footnote{For every equation except the Allen-Cahn equation, we use an upper limit of $2,000$ fine-tuning epochs.
In these experiments, the $L^p$ loss generally stops changing by that time. 
For the Allen-Cahn equation, however, the $L^p$ loss takes much longer to stabilize, so we use an upper limit of $10,000$ fine-tuning epochs.} of fine-tuning epochs are complete.

\bigskip

In our experiments, all three steps use the Adam optimizer \cite{kingma2014adam} with a learning rate of $10^{-3}$.
Though we do not use it in our experiments, our implementation supports the LBFGS optimizer \cite{liu1989limited}.

\bigskip

In every experiment in this section, we set $p = 0.1$ and $N_{Random\ Coll} = 3,000$ (see section \ref{sub_sec:Training}).
We did not attempt to optimize these values and do not claim they are optimal.
However, we found them to be sufficient in our experiments.

\begin{figure}[!hbt]
    \centering
    \includegraphics[width=.8\linewidth]{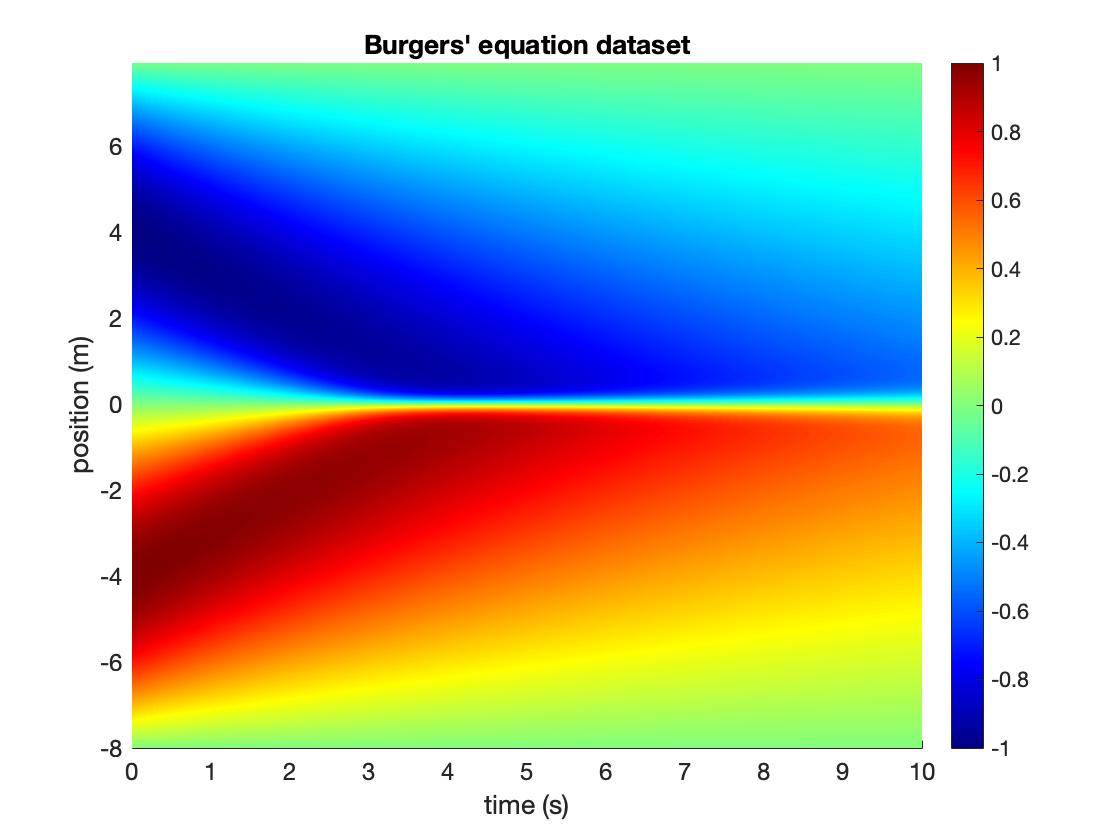}
    \caption{Noise-free Burgers' equation data set.}
    \label{fig:Burgers_Dataset}
\end{figure}

\subsection{Burgers' equation} \label{sub_sec:Burgers}
Burgers' equation is a non-linear second-order non-linear PDE.
It was first studied in \cite{bateman1915some} but arises in many contexts, including Fluid Mechanics, Nonlinear Acoustics, Gas Dynamics, and Traffic Flow \cite{basdevant1986spectral}. 
Burgers' equation has the form

\begin{equation} D_{t} u = \nu (D_{x}^2 u) - (u) (D_{x} u), \label{eq:Burgers} \end{equation}

where velocity, $u$, is a function of $t$ and $x$.
Here, $\nu > 0$ is the {\it diffusion coefficient}.
Significantly, solutions to Burgers' equation can develop shocks (discontinuities).

\bigskip 

We test \texttt{PDE-LEARN} on a Burgers' equation with $\nu = 0.1$ on the domain $(t, x) \in [0, 10] \times [-8, 8]$. 
Thus, for these experiments, $n_S = 1$.
For this data set,

$$u(0, x) = -\sin\left( \frac{\pi x}{8} \right).$$

To make the noise-free data set, we partition the problem domain using a spatiotemporal grid with $257$ grid lines along the $x$-axis and $201$ along the $t$-axis. 
Thus, each grid square has a length of $1/16$ along the $x$-axis and a length of $0.05$ along the $t$-axis. 
Our script \texttt{Burgers\_Sine.m} (in the \texttt{MATLAB} sub-directory of our repository) uses \texttt{Chebfun}'s \cite{driscoll2014chebfun} \texttt{spin} class to find a numerical solution to Burgers' equation on this grid.
Figure \ref{fig:Burgers_Dataset} depicts the noise-free data set. 

\bigskip 

Using the procedure outlined at the beginning of section \ref{sec:Experiments}, we generate several noisy, limited data sets from the noise-free data set.
We test \texttt{PDE-LEARN} on each data set.
In each experiment, $U$ contains five layers with $20$ neurons per layer. 
For these experiments, the left-hand side term is

$$f_0 \Big(\hat{\partial}^0 u, \ldots, \hat{\partial}^{N_M} u \Big) = D_t U.$$

Likewise, the right-hand side terms are

\begin{align*} U, D_x U, &D_x^2 U, D_x^3 U, \\
(U)^2, (D_x U)U, &(D_x^2 U)U, (D_x U)^2, \\
(U)^3, (D_x U)(U)^2, &(D_x^2 U) U^2, (D_x U)^2 U , \\
(U)^4, (D_x U)(U)^3, &(D_x^2 U) (U)^3, (D_x U)^2(U)^2, (D_x U)^3 U \end{align*}

Thus, our library includes terms with up to third-order spatial derivatives and fourth-order multiplicative products.
For these experiments, we use $1,000$ burn-in epochs ($N_{Burn-in}$), $1,000$ sparsification epochs ($N_{Sparse}$) with $w_{L^p} = 0.0001$, and a variable number of fine-tuning epochs ($N_{Fine-tune}$). 
Table \ref{Table:Experiments:Burgers_Sine} reports the results of our experiments with Burgers' equation.

\bigskip

\begin{table}[hbt]
    \centering 
    \rowcolors{2}{cyan!10}{white}
    \begin{threeparttable}
        \caption{Experimental results for Burgers' equation} 
        \label{Table:Experiments:Burgers_Sine}

        \begin{tabulary}{0.9\linewidth}{p{0.7cm}p{0.6cm}p{1.3cm}p{1.0cm}p{1.5cm}L}
            \toprule[0.3ex]
            $N_{Data}$ & Noise & $N_{Burn-in}$ & $N_{Sparse}$ & $N_{Fine-tune}$ & Identified PDE \\
            \midrule[0.1ex] 
            $4,000$ & $50\%$ & $1,000$ & $1,000$ & $2,000$ & $D_t U = 0.0987(D_x^2 U) - 0.9704(D_x U)(U)$ \\
            \addlinespace[0.4em]
            $4,000$ & $75\%$ & $1,000$ & $1,000$ & $1,000$ & $D_t U = 0.1025(D_x^2 U) -  0.9883(D_x U)(U)$ \\
            \addlinespace[0.4em]
            $4,000$ & $100\%$ & $1,000$ & $1,000$ & $400$ & $D_t U = 0.0824(D_x^2 U)  - 0.8476(D_x U)(U)$ \\
            \addlinespace[0.4em]
            $2,000$ & $50\%$ & $1,000$ & $1,000$ & $1,000$ & $D_t U =  0.0850(D_x^2 U) -  0.9010(D_x U)(U)$ \\
            \addlinespace[0.4em]
            $2,000$ & $75\%$ & $1,000$ & $1,000$ & $0$ & $D_t U =  0.0608(D_x^2 U) -  0.7067(D_x U)(U)$ \\
            \addlinespace[0.4em]
            $2,000$ & $100\%$ & $1,000$ & $1,000$ & $750$ & $D_t U =  0.0582(D_x^2 U) - 0.6756(D_x U)(U)$ \\
            \addlinespace[0.4em]
            $1,000$ & $25\%$ & $1,000$ & $1,000$ & $1,000$ & $D_t U =  0.1003(D_x^2 U) - 0.9880(D_x U)(U)$ \\
            \addlinespace[0.4em]
            $1,000$ & $50\%$ & $1,000$ & $1,000$ & $200$ & $D_t U = 0.0839(D_x^2 U) - 0.7977(D_x U)(U)$ \\
            \addlinespace[0.4em]
            $1,000$ & $75\%$\tnote{1} & $1,000$ & $1,000$ & $0$ & $D_t U =  0.0386(D_x^2 U) -  0.5327(D_x U)(U)$ \\
            \addlinespace[0.4em]
            $500$ & $10\%$ & $1,000$ & $1,000$ & $1,000$ & $D_t U =  0.1002(D_x^2 U) - 0.9723(D_x U)(U)$ \\
            \addlinespace[0.4em]
            $500$ & $25\%$ & $1,000$ & $1,000$ & $1,000$ & $D_t U =  0.0876(D_x^2 U) - 0.9714(D_x U)(U)$\\
            \addlinespace[0.4em]
            $500$ & $50\%$\tnote{1} & $1,000$ & $1,000$ & $0$ & $D_t U =  0.0394(D_x^2 U) - 0.7171(D_x U)(U)$ \\    
            \addlinespace[0.4em]
            $250$ & $10\%$ & $1,000$ & $1,000$ & $200$ & $D_t U =  - 0.0638(U) -  0.0969(D_x U)(U)$ \\
            \addlinespace[0.4em]
            $250$ & $25\%$ & $1,000$ & $1,000$ & $1,000$ & $D_t U =  0.0948(D_x^2 U) - 0.8346(D_x U)(U)$ \\
            \bottomrule[0.3ex]
        \end{tabulary}

        \begin{tablenotes}
            \item [1] For these experiment, we use $W_{L^p} = 0.0002$ during the sparsification step. If we set $w_{L^p} = 0.0001$, \texttt{PDE-LEARN} identifies $D_t U = - 0.0373(U) + 0.0385(D_x^2 U) - 0.5168(D_x U)(U)$ in the $N_{Data} = 1,000$, $75\%$ noise experiment and $D_t U = -0.0379(U) + 0.0327(D_x^2 U) -0.6407(D_x U)*(U) -0.0132(D_x U)^2$ in the $N_{Data} = 500$, $50\%$ noise experiment.
        \end{tablenotes}
    \end{threeparttable}
\end{table}

Thus, \texttt{PDE-LEARN} successfully learns Burgers' equation in all but one of the experiments.
\texttt{PDE-LEARN} can identify Burgers' equation from $2,000$ data points even when we corrupt the data set with $100\%$ noise.
If the noise level decreases to just $50\%$, \texttt{PDE-LEARN} can reliably identify Burgers' equation with as few as $500$ data points.
Increasing the noise tends to increase the relative error between the identified and the corresponding coefficients in the hidden PDE.
With that said, in the $75\%$ noise and $4,000$ data point experiment, the relative error of identified coefficients is $\approx 1\%$.

\bigskip

Notably, \texttt{PDE-LEARN} fails to identify Burgers' equation in one of the $N_{Data} = 250$ experiments.
Even in this experiment, however, \texttt{PDE-LEARN} correctly identifies the RHS term $(D_x U)(U)$.
This result suggests that even when PDE-LEARN fails, it may still extract useful information about the hidden PDE.
Interestingly, when $N_{Data} = 250$, \texttt{PDE-LEARN} successfully identifies Burgers’ equation when the noise level is $25\%$ but not when it is $10\%$. 
This result suggests that the number of measurements, not the noise level, is the main limiting factor in the low-data limit.

\subsection{KdV Equation} \label{sub_sec:KdV}

Next, we consider The Korteweg–De Vries (KdV) equation, a non-linear third-order equation.
\cite{korteweg1895xli} derived the KdV equation to describe the evolution of one-dimensional, shallow-water waves.
With appropriate scaling, the KdV equation is

\begin{equation} D_{t} u = -(u)(D_{x} u) - D_{x}^3 u, \label{eq:KdV} \end{equation}

where wave height, $u$, is a function of $x$ and $t$. 

\bigskip

We test \texttt{PDE-LEARN} on the KdV equation on the domain $(t, x) \in [0, 40] \times [-20, 20]$. 
For this equation, we consider two initial conditions:

$$u(0, x) = -\sin\left( \frac{\pi x}{20} \right),$$

and 

$$u(0, x) = \exp\left(-\pi\left(\frac{x}{30}\right)^2\right) \cos\left(\frac{\pi x}{10} \right).$$

\begin{figure}[!hbt]
    \centering
    \includegraphics[width=.8\linewidth]{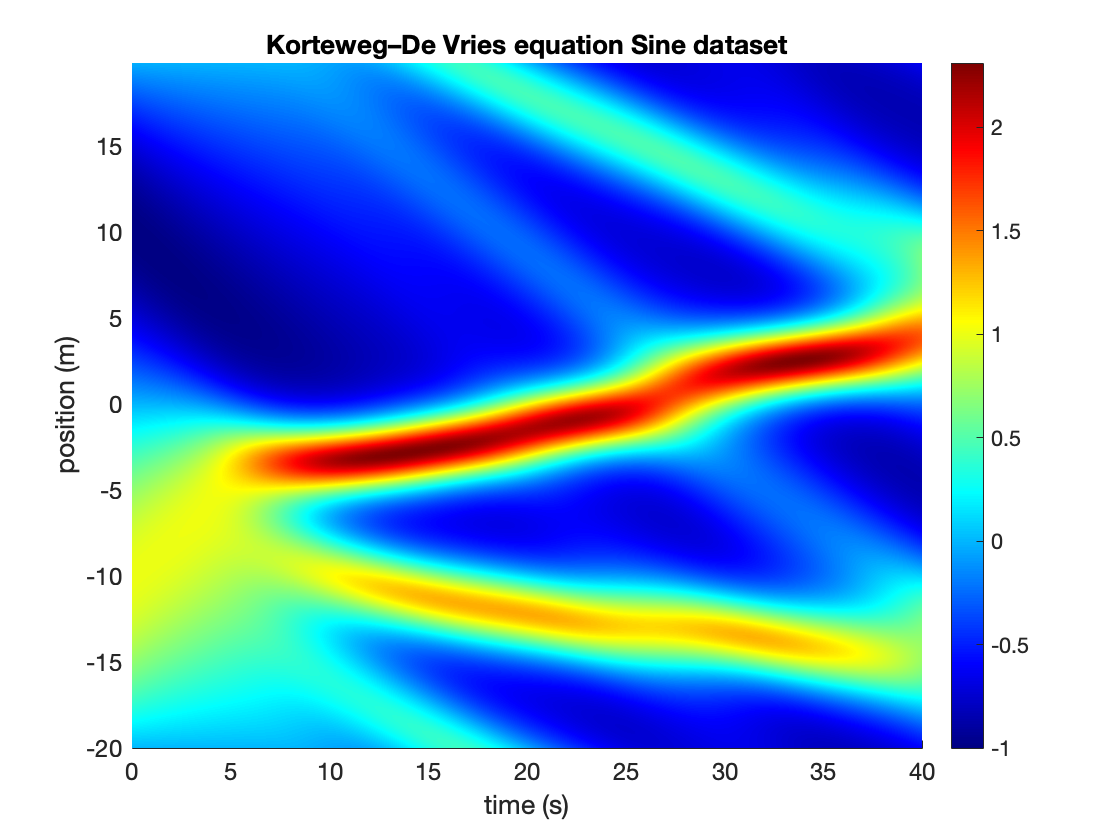}
    \caption{Noise-free KdV-$\sin$ data set.}
    \label{fig:KdV_Sine_Dataset}
\end{figure}

\begin{figure}[!hbt]
    \centering
    \includegraphics[width=0.8\linewidth]{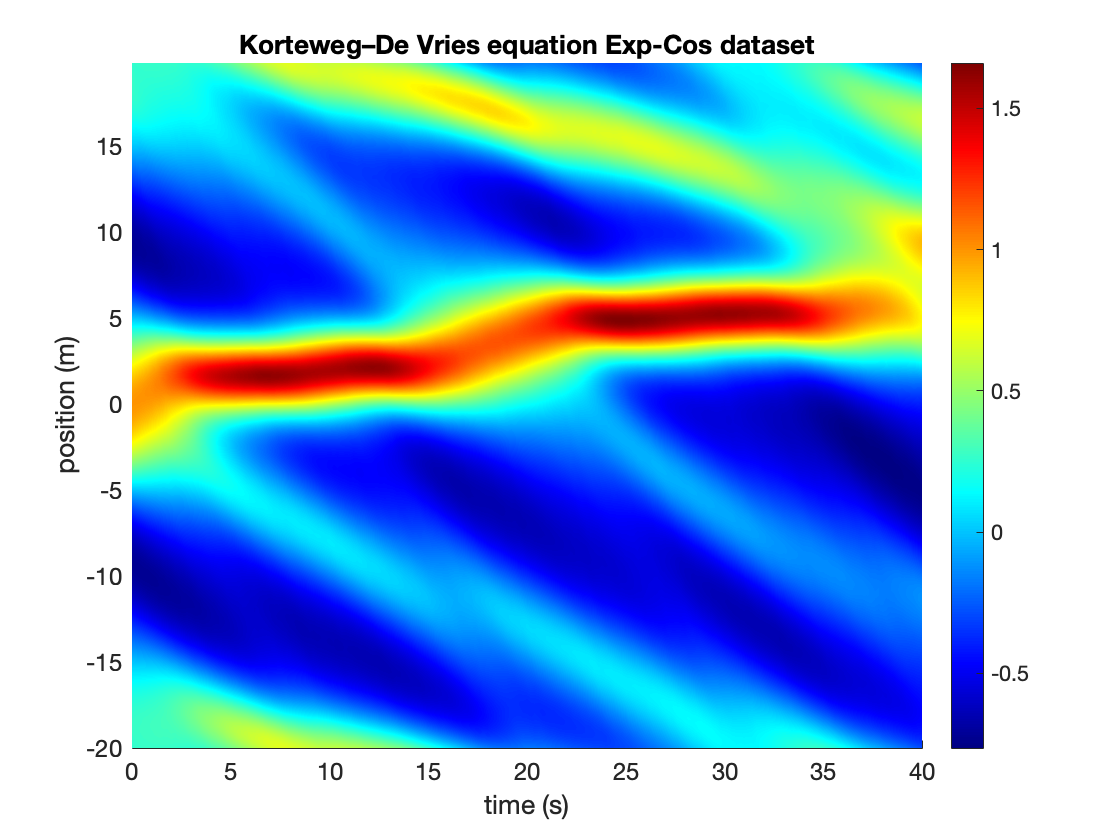}
    \caption{Noise-free KdV-$\exp$-$\cos$ data set.}
    \label{fig:KdV_Exp_Cos_Dataset}        
\end{figure}

We partition the problem domain into a spatiotemporal grid with $257$ grid lines along the $x$-axis and $201$ along the $t$-axis. 
We use \texttt{Chebfun}'s \cite{driscoll2014chebfun} \texttt{spin} class to solve the KdV equation with each initial condition on this grid. 
We refer to the resulting solutions as our noise-free KdV-$\sin$ and KdV-$\exp$-$\cos$ data sets, respectively.
The scripts \texttt{KdV\_Sine.m} and \texttt{KdV\_Exp\_Cos.m} in the \texttt{MATLAB} sub-directory of our repository generate these data sets.
Figures \ref{fig:KdV_Sine_Dataset} and \ref{fig:KdV_Exp_Cos_Dataset} depict the data sets. 

\bigskip 

In all of our experiments with the KdV equation, $U$ contains four layers with $40$ neurons per layer\footnote{We tried using the same architecture as in the Burgers’ experiments.
However, we found that architecture was too simple to learn the intricacies of the KdV data sets.}.
Further, we use the same left and right-hand side terms as in the Burgers' experiments.

\bigskip

\subsubsection{KdV sin data set}
\label{sub_sub_sec:KdV:Sine}
We test \texttt{PDE-LEARN} on several noisy, limited data sets built using the two noise-free data sets.
We test \texttt{PDE-LEARN} on each data set.
For these experiments, burn-in takes between $1,000$ and $2,000$ epochs using the \texttt{Adam} optimizer (stopping once the data loss stops decreasing).
For the sparsification step, we use $w_{L^p} = 0.0002$ and train for between $1,000$ and $2,000$ epochs (stopping once the $L^p$ loss remains roughly constant for at least $200$ epochs). 
Table \ref{Table:Experiments:KdV_Sine} reports our experimental results with the KdV-$\sin$ data set. 

\begin{table}[hbt]
    \centering 
    \rowcolors{2}{cyan!10}{white}
    \begin{threeparttable}
        \caption{Experimental results for the KdV-$\sin$ data set} 
        \label{Table:Experiments:KdV_Sine}

        \begin{tabulary}{.9\linewidth}{p{0.7cm}p{0.6cm}p{1.3cm}p{1.0cm}p{1.5cm}L}
            \toprule[0.3ex]
            $N_{Data}$ & Noise & $N_{Burn-in}$ & $N_{Sparse}$ & $N_{Fine-tune}$ & Identified PDE \\
            \midrule[0.1ex]
            $4,000$ & $50\%$ & $2,000$ & $2,000$ & $1,000$ & $D_t U = -0.8010(D_x^3 U) - 0.8339(D_x U)(U)$ \\
            \addlinespace[0.4em]
            $4,000$ & $75\%$ & $1,000$ & $1,000$ & $300$ & $D_t U =  -0.5650(D_x^3 U) - 0.5954(D_x U)(U)$ \\
            \addlinespace[0.4em]
            $2,000$ & $25\%$ & $1,000$ & $1,000$ & $2,000$ & $D_t U = -0.9101(D_x^3 U) -  0.9217(D_x U)(U)$ \\
            \addlinespace[0.4em]
            $2,000$ & $50\%$ & $1,500$ & $2,000$ & $2,000$ & $D_t U =  -  0.8680(D_x^3 U) -  0.8778(D_x U)(U)$ \\
            \addlinespace[0.4em]
            $1,000$ & $10\%$ & $1,000$ & $1,000$ & $2,000$ & $D_t U =  -  0.9249(D_x^3 U) -  0.9245(D_x U)(U)$ \\
            \addlinespace[0.4em]
            $1,000$ & $25\%$ & $1,500$ & $1,500$ & $2,000$ & $D_t U =  -  0.8752(D_x^3 U) -  0.8937(D_x U)(U)$ \\
            \addlinespace[0.4em]
            $1,000$ & $50\%$ & $1,000$ & $1,000$ & $100$ & $D_t U =  -0.0889(D_x U)(U)$ \\
            \addlinespace[0.4em]
            $500$ & $10\%$ & $1,000$ & $1,000$ & $2,000$ & $D_t U =  -  0.9183(D_x^3 U) -  0.9311(D_x U)(U)$\\
            \addlinespace[0.4em]
            $500$ & $25\%$ & $1,000$ & $1,500$ & $500$ & $D_t U =  -  0.6641(D_x^3 U) -  0.7350(D_x U)(U)$ \\
            \addlinespace[0.4em]
            $500$ & $50\%$ & $1,000$ & $1,000$ & $0$ & $D_t U =  -0.0947(D_x U)(U)$ \\    
            \addlinespace[0.4em]
            $250$ & $10\%$ & $1,000$ & $1,000$ & $0$ & $D_t U =  -  0.4514(D_x^3 U) - 0.4400(D_x U)*(U) - 0.1034(D_x U)^3(U)$ \\
            \addlinespace[0.4em]
            $250$ & $25\%$ & $1,000$ & $1,000$ & $0$ & $D_t U =  -  0.0943(D_x^3 U) - 0.0725(D_x U)*(U) - 0.1347(D_x U)^3(U)$ \\
            \bottomrule[0.3ex]
        \end{tabulary}

    \end{threeparttable}
\end{table}

\bigskip

These results show that \texttt{PDE-LEARN} can identify the KdV equation from limited measurements, even at high noise levels. 
As in the Burgers experiments, the identified coefficients tend to be more accurate in lower noise experiments.
Interestingly, \texttt{PDE-LEARN} can identify the KdV equation from the $\sin$ data set with as few as $500$ data points and $25$\% noise.

\bigskip

With that said, \texttt{PDE-LEARN} does have its limits.
In particular, \texttt{PDE-LEARN} fails to identify the KdV equation in a few experiments.
The identified PDE is too sparse in the $1,000$ data points, $50$\% noise, and $500$ data points, $50$\% noise experiments.
In both cases, however, \texttt{PDE-LEARN} correctly identifies the RHS-term $(D_x U)(U)$.
By contrast, the identified PDE contains extra terms in the $250$ data points, $10$\% noise, and $250\%$ data points, $25\%$ noise experiments.
In these experiments, the identified PDE contains both terms of the KdV equation (in addition to some extra ones). 
These results strengthen our assertion that \texttt{PDE-LEARN} identifies useful information even when it fails.
If the identified PDE is too sparse, the terms in the identified PDE are likely to be present in the hidden PDE.
Likewise, if the identified is not sparse enough, the terms of the hidden PDE are likely to be in the identified PDE.

\bigskip

\subsubsection{KdV exp-cos data set}
\label{sub_sub_sec:KdV:Exp_Cos}
Next, we test \texttt{PDE-LEARN} on the KdV $\exp$-$\cos$ data set.
For these experiments, burn-in takes $1,000$ epochs. 
For the sparsification step, we set $w_{L^p} = 0.0001$ and train for $1,000$ epochs.
Table \ref{Table:Experiments:KdV_Exp_Cos} reports our experimental results with the KdV $\exp$-$\cos$ data set. 

\bigskip

\begin{table}[hbt]
    \centering 
    \rowcolors{2}{cyan!10}{white}
    \begin{threeparttable}
        \caption{Experimental results for the KdV-$\exp$-$\cos$ data set} 
        \label{Table:Experiments:KdV_Exp_Cos}

        \begin{tabulary}{.9\linewidth}{p{0.7cm}p{0.6cm}p{1.3cm}p{1.0cm}p{1.5cm}L}
            \toprule[0.3ex]
            $N_{Data}$ & Noise & $N_{Burn-in}$ & $N_{Sparse}$ & $N_{Fine-tune}$ & Identified PDE \\
            \midrule[0.1ex]
            $4,000$ & $50\%$ & $1,000$ & $1,000$ & $2,000$ & $D_t U =  -0.8302(D_x^3 U) - 0.8700(D_x U)(U))$ \\
            \addlinespace[0.4em]
            $4,000$ & $75\%$ & $1,000$ & $1,000$ & $300$ & $D_t U =  -0.5758(D_x^3 U) - 0.6384(D_x U)(U)$ \\
            \addlinespace[0.4em]
            $2,000$ & $25\%$ & $1,000$ & $1,000$ & $2,000$ & $D_t U = -0.9101(D_x^3 U) -  0.9217(D_x U)(U)$ \\
            \addlinespace[0.4em]
            $2,000$ & $50\%$ & $1,000$ & $1,000$ & $1,000$ & $D_t U =  - 0.5585(D_x^3 U) -  0.6066(D_x U)(U)$ \\
            \addlinespace[0.4em]
            $1,000$ & $10\%$ & $1,000$ & $1,000$ & $2,000$ & $D_t U =  - 0.9145(D_x^3 U) -  0.9229(D_x U)(U)$ \\
            \addlinespace[0.4em]
            $1,000$ & $25\%$ & $1,000$ & $1,000$ & $2,000$ & $D_t U =  -  0.7976(D_x^3 U) -  0.8157(D_x U)(U)$ \\
            \addlinespace[0.4em]
            $1,000$ & $50\%$ & $600$\tnote{1} & $1,000$ & $1,000$ & $D_t U =  -  0.1006(D_x U)(U)$ \\
            \addlinespace[0.4em]
            $500$ & $10\%$ & $1,000$ & $1,000$ & $2,000$ & $D_t U =  -  0.9698(D_x^3 U) -  0.9322(D_x U)(U)$\\
            \addlinespace[0.4em]
            $500$ & $25\%$ & $1,000$ & $1,000$ & $200$ & $D_t U =  -  0.1303(D_x U)(U) -  0.0584(D_x U)^3(U)$ \\
            \addlinespace[0.4em]
            $500$ & $50\%$ & $1,000$ & $1,000$ & $700$ & $D_t U = 0.1323(D_x U) -  0.1818(D_x U)(U)$ \\    
            \addlinespace[0.4em]
            $250$ & $10\%$ & $1,000$ & $1,000$ & $500$ & $D_t U =  -  0.7927(D_x^3 U) -  0.8440(D_x U)(U)$ \\
            \addlinespace[0.4em]
            $250$ & $25\%$ & $1,000$ & $1,000$ & $200$ & $D_t U =  -  0.1028(D_x^3 U) -  0.2467(D_x U)(U)$ \\
            \bottomrule[0.3ex]
        \end{tabulary}

        \begin{tablenotes}
            \item [1] We stop the burn in step after just $600$ epochs for the $1000$ data point, $50$\% noise experiment. 
            This is because the solution network began over-fitting the data set.
        \end{tablenotes}

    \end{threeparttable}
\end{table}

Once again, \texttt{PDE-LEARN} correctly identified the KdV equation under several noise levels across several data set sizes. 
Significantly, it identifies the KdV equation from just $250$ data points with $25$\% noise. 
Further, as in with the $\sin$ data set, \texttt{PDE-LEARN} successfully identified the KdV equation from $4,000$ data points and $75$\% noise.

\bigskip 

As with the $\sin$ data set, however, \texttt{PDE-LEARN} does have limits.
It fails to identify Burger's equation in the $1,000$ data points, $50$\% noise experiment, and the $500$ data points, $50$\% noise experiments.
In the former, the identified PDE is too sparse, though the lone term in the identified PDE, $(D_x U)(U)$, is present in the KdV equation.
The latter is more concerning, as the term $D_x U$ is in the identified PDE, while one of the terms of the KdV equation, $D_x^3 U$, is not. 
With that said, even in this case, \texttt{PDE-LEARN} did correctly identify one of the terms in the KdV equation, $(D_x U)(U)$. 
These result suggest that \texttt{PDE-LEARN} fails only under extreme conditions and can still yield useful information when it does fail.

\bigskip

\subsubsection{Combined data sets}
\label{sub_sub_sec:KdV:Combined}
Finally, to demonstrate that \texttt{PDE-LEARN} can learn from multiple data sets simultaneously (assuming the corresponding system response functions satisfy a common PDE), we test with the $\sin$ and $\exp$-$\cos$ data sets simultaneously. 
For these experiments, burn-in takes $1,000$ epochs using the \texttt{Adam} optimizer (stopping once the data loss stops decreasing).
For the sparsification step, we use $w_{L^p} = 0.0001$ and train for between $1,000$ epochs.
For the combined data sets, we only consider conditions that cause \texttt{PDE-LEARN} trouble when learning from a single data set. 
Table \ref{Table:Experiments:KdV_Combined} reports our experimental results for the combined KdV data set experiments.

\bigskip 

\begin{table}[hbt]
    \centering 
    \rowcolors{2}{white}{cyan!10}
    \begin{threeparttable}
        \caption{Experimental results for the combined KdV data sets} 
        \label{Table:Experiments:KdV_Combined}

        \begin{tabulary}{.9\linewidth}{p{0.7cm}p{0.6cm}p{1.3cm}p{1.0cm}p{1.5cm}L}
            \toprule[0.3ex]
            $N_{Data}$ & Noise & $N_{Burn-in}$ & $N_{Sparse}$ & $N_{Fine-tune}$ & Identified PDE \\
            \midrule[0.1ex]
            $1,000$ & $25\%$ & $1,000$ & $1,000$ & $2,000$ & $(D_t U) =  -0.0233(D_x^3 U) -  0.1286(D_x U)(U)$ \\
            \addlinespace[0.4em]
            $1,000$ & $50\%$ & $1,000$ & $1,000$ & $0$ & $(D_t U) =  -0.8475(D_x^3 U) -  0.8719(D_x U)(U)$ \\
            \addlinespace[0.4em]
            $500$ & $25\%$ & $1,000$ & $1,000$ & $1,000$ & $(D_t U) =  -0.6778(D_x^3 U) -  0.7282(D_x U)(U)$ \\
            \addlinespace[0.4em]
            $500$ & $50\%$ & $500$ & $1,000$ & $0$ & $(D_t U) = -0.0794(D_x U)(U)$ \\
            \addlinespace[0.4em]
            $25$ & $25\%$ & $1,000$ & $1,000$ & $0$ & $(D_t U) = - 0.1267(D_x U)(U)$ \\
            \bottomrule[0.3ex]
        \end{tabulary}

        \begin{tablenotes}
            \item [1] Due to over-fitting, we stop burn in after $600$ epochs in the $1000$ data point, $50$\% noise experiment. 
        \end{tablenotes}

    \end{threeparttable}
\end{table}

Notably, \texttt{PDE-LEARN} successfully identifies the KdV equation when each data set contains $1000$ data points and $50$\% noise, even though it can not identify the KdV equation from either data set individually. 
Further, in the experiments we ran on both the individual and the combined data sets, the identified coefficients tend to be more accurate in the combined experiments.
These results suggest that using multiple data sets can improve \texttt{PDE-LEARN}'s performance.

\bigskip

Even with the combined data set, \texttt{PDE-LEARN} has limitations.
It fails to identify the KdV equation in the $500$ data points, $50$\% noise, and the $250$ data points, $25$\% noise experiments. 
As in previous experiments, even when \texttt{PDE-LEARN} fails, the terms in the identified PDEs are present in the KdV equation.
Thus, \texttt{PDE-LEARN} can still uncover useful information, even when it can not identify the hidden PDE.

\bigskip

Our experiments with the KdV equation suggest that \texttt{PDE-LEARN}’s performance degrades when using fewer than $500$ data points, irrespective of the noise level. 
Above this threshold, however, \texttt{PDE-LEARN} appears to be reliable, even in the presence of significant noise.

\begin{figure}[!hbt]
    \centering
    \includegraphics[width=0.8\linewidth]{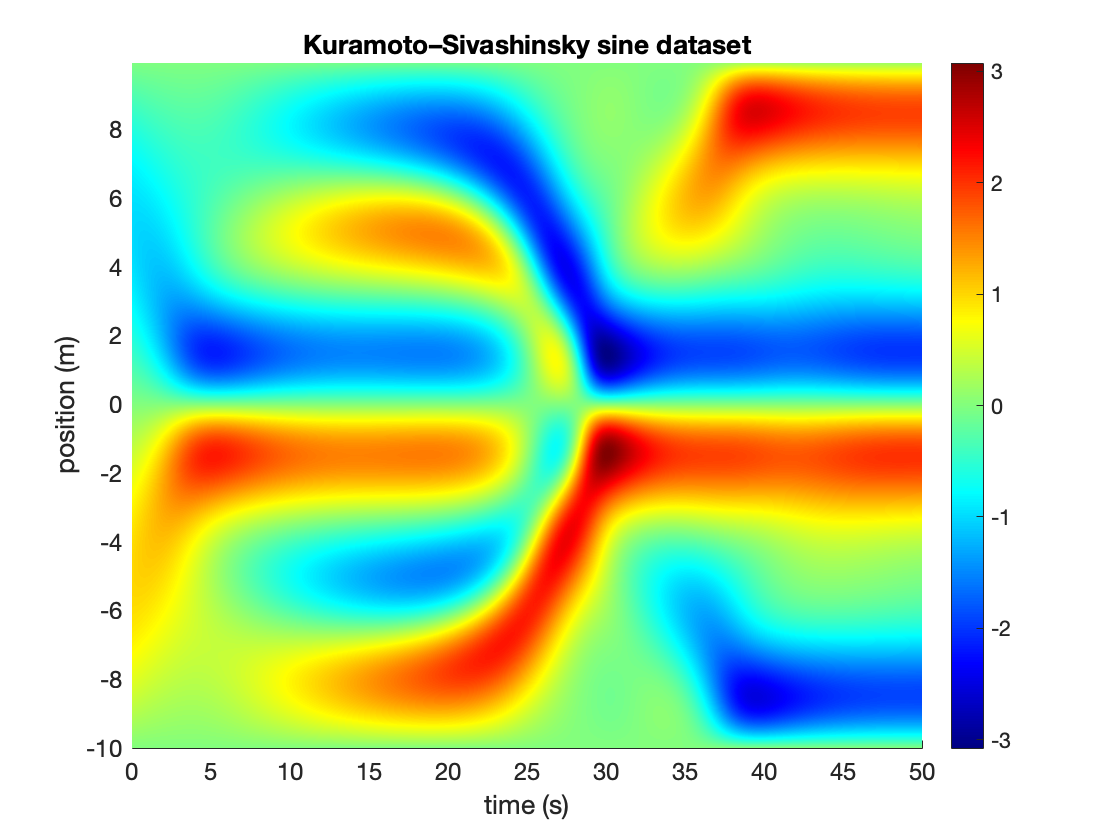}
    \caption{Noise-free KS equation data set.}
    \label{fig:KS_Sine_Dataset}        
\end{figure}

\subsection{Kuramoto–Sivashinsky equation}
\label{sub_sec:KS}

The Kuramoto–Sivashinsky (KS) equation \cite{kuramoto1976persistent} \cite{sivashinsky1977nonlinear} is a non-linear fourth-order equation that arises in many physical contexts, including flame propagation, plasma physics, chemical physics, and combustion dynamics \cite{papageorgiou1991route}.
In one dimension, the KS equation takes the following form:

\begin{equation} D_t u = \nu D_x^2 u - \mu D_x^4 u - \lambda (u)(D_x u) \label{eq:KS} \end{equation}

If $\nu < 0$, solutions to the KS equation can be chaotic with violent shocks \cite{kuramoto1976persistent} \cite{papageorgiou1991route}.

\bigskip

We test \texttt{PDE-LEARN} on the KS equation with $\nu = -1$, $\mu = 1$, and $\lambda = 1$.
For these experiments, our problem domain is $(t, x) \in (0, T] \times S = (0, 5] \times (-5, 5)$. 
We partition $S = (-5, 5)$ into $255$ equally-sized sub-intervals and $(0, T] = (0, 5]$ into $200$ equally-sized sub-intervals.
This partition engenders a regular grid with $256$ equally-spaced grid lines along the $x$-axis and $201$ along the $t$-axis.
We use \texttt{Chebfun}'s \texttt{spin} class to solve the KS equation on this grid subject to the initial condition 

$$ U(0, x) = \cos \left( \frac{2 \pi x}{5} \right)\left( 1 + \sin \left( \frac{\pi x}{5} \right) \right). $$

The resulting solution is our noise-free $\sin$ data set, which we depict in figure \ref{fig:KS_Sine_Dataset}.

\bigskip 

Using the procedure outlined at the beginning of section \ref{sec:Experiments}, we generate several noisy, limited data sets from the noise-free data set.
We test \texttt{PDE-LEARN} on each data set.
In each experiment, $U$ contains four layers with $40$ neurons per layer. 
Further, we use the same left-hand and right-hand side terms as in the previous experiments, except that we add $D_x^4 U$ to the RHS terms.
For these experiments, we use $2,000$ burn-in epochs ($N_{Burn-in}$), $1,000 - 2,000$ sparsification epochs ($N_{Sparse}$) with $w_{L^p} = 0.0003$, and a variable number of fine-tuning epochs ($N_{Fine-tune}$). 
Further, during the sparsification step, we set $w_{L^p} = 0.0003$.
Table \ref{Table:Experiments:KS_Sine} reports the results of our experiments with the KS equation.

\bigskip

\begin{table}[hbt]
    \centering 
    \rowcolors{2}{cyan!10}{white}
    \begin{threeparttable}
        \caption{Experimental results for the KS $\sin$ data set} 
        \label{Table:Experiments:KS_Sine}

        \begin{tabulary}{.9\linewidth}{p{0.7cm}p{0.6cm}p{1.3cm}p{1.0cm}p{1.5cm}L}
            \toprule[0.3ex]
            $N_{Data}$ & Noise & $N_{Burn-in}$ & $N_{Sparse}$ & $N_{Fine-tune}$ & Identified PDE \\
            \midrule[0.1ex]
            $4,000$ & $5\%$ & $2,000$ & $1,000$ & $2,000$ & $D_t U =  -  0.8118(D_x^2 U) -  0.8202(D_x^4 U) -  0.8579(D_x U)(U)$ \\
            \addlinespace[0.4em]
            $4,000$ & $10\%$ & $2,000$ & $1,000$ & $2,000$ & $D_t U =  -  0.7747(D_x^2 U) -  0.7895(D_x^4 U) -  0.8252(D_x U)(U)$ \\
            \addlinespace[0.4em]
            $4,000$ & $15\%$ & $2,000$ & $2,000$ & $1,000$ & $D_t U =  -  0.6875(D_x^2 U) -  0.7067(D_x^4 U) -  0.7662(D_x U)(U)$ \\
            \addlinespace[0.4em]
            $4,000$ & $20\%$ & $2,000$ & $1,000$ & $500$ & $D_t U =  0.1395(U) +  0.1808(D_x^2 U) -  0.4439(D_x U)(U) +  0.0592(D_x U)*(U)^3 +  0.0834(D_x U)^3 (U)$ \\
            \bottomrule[0.3ex]
        \end{tabulary}
    \end{threeparttable}
\end{table}

Thus, \texttt{PDE-LEARN} can successfully identify the KS equation with up to $15\%$ noise.
In the $20\%$ noise experiment, the identified PDE does not contain the $D_x^4 U$ term but does contain terms that are not present in the KS equation.
This result is a notable departure from the results we observe with other equations, where misidentified PDEs contain too many or too few terms but never both.
Even in this case, however, the identified PDE contains two of the terms of the KS equation. 
Thus, PDE-LEARN still recovers useful information.
These results suggest that while \texttt{PDE-LEARN} can identify the KS equation, it appears to be less robust with this equation than with other equations we consider in this section.
We discuss this result in section \ref{sub_sec:Discussion:Limits}.

\begin{figure}[!hbt]
    \centering
    \includegraphics[width=0.8\linewidth]{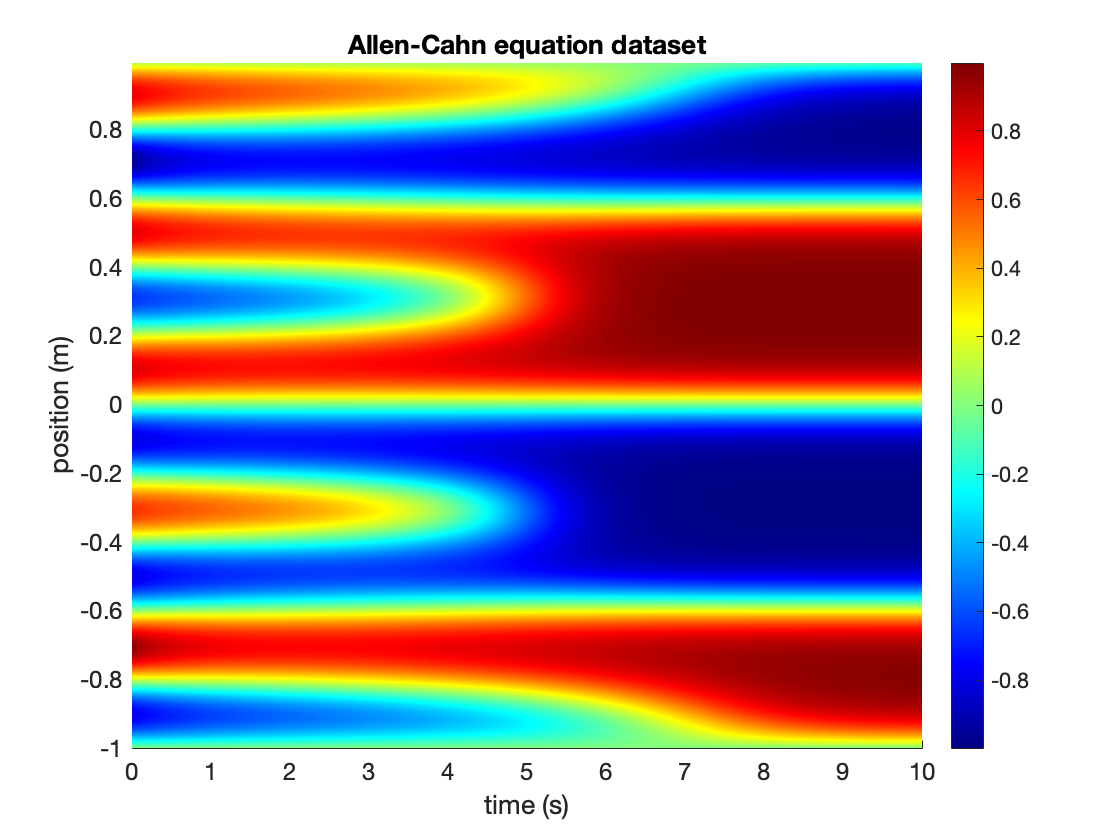}
    \caption{Noise-free Allen-Cahn equation data set.}
    \label{fig:Allen_Cahn_Dataset}        
\end{figure}

\subsection{Allen-Cahn equation}
\label{sec:Allen-Cahn}

Next, we consider the Allen-Cahn equation. 
This second-order non-linear PDE describes the separation of the constituent metals in of multi-component molten alloy mixture
\cite{allen1979microscopic}.
The Allen-Cahn equation has the following form:

\begin{equation} D_{t}u = \varepsilon (D_{x}^2 u) - u^3 + u. \label{eq:Allen-Cahn} \end{equation}

\bigskip

We test \texttt{PDE-LEARN} on the Allen-Cahn equation with $\varepsilon = 0.003$ on the domain $(t, x) \in [0, 40] \times [-20, 20]$ with the initial condition

$$u(0, x) = -0.2 \sin\left(2 \pi x\right)^{5} + 0.8 \sin\left(5\pi x\right).$$

The Allen-Cahn equation with this value of $\varepsilon$ presents an interesting challenge for $\texttt{PDE-LEARN}$ because it contains a small coefficient.
As described in section \ref{sec:Methodology}, \texttt{PDE-LEARN} eliminates all components of $\xi$ which are smaller than a pre-defined threshold.
In all our e.pngxperiments\footnote{Notably, we could rewrite \texttt{PDE-LEARN} to use double-precision floating point numbers, allowing for a much smaller threshold. 
We discuss this in section 6.3.}, the threshold is around $10^{-4}$.
As such, the Allen-Cahn equation contains a coefficient close to the threshold.
Therefore, if the identified coefficient is too far off from the true value, it risks being thresholded. 
Notwithstanding, \texttt{PDE-LEARN} performs admirably, even when the hidden PDE contains coefficients that are close to the threshold. 

\bigskip 

We partition the problem domain into a spatiotemporal grid with $257$ equally-spaced grid lines along the $x$-axis and $201$ along the $t$-axis. 
We use \texttt{Chebfun}'s  \cite{driscoll2014chebfun} \texttt{spin} class to solve the Allen-Cahn equation on this grid. 
We refer to the resulting solutions as our noise-free Allen-Cahn data set.
The script \texttt{Allen\_Cahn.m} in the \texttt{MATLAB} sub-directory of our repository generates this data set.
Figure \ref{fig:Allen_Cahn_Dataset} depicts the data sets. 

\bigskip 

For these experiments, $U$ contains five layers with $20$ neurons per layer.
Further, we use the same left-hand and right-hand side terms as in the Burgers' and KdV experiments.
Table \ref{Table:Experiments:Allen_Cahn} reports the results of our Allen-Cahn equation experiments.
In each experiment, the burn-in step lasts for $2,000$ epochs ($N_{Burn-in} = 2,000$), the sparsification step lasts $1,000$ epochs ($N_{Sparse} = 1,000$), and the fine-tuning step lasts for $10,000$ epochs ($N_{Fine-tune} = 10,000$). 
Further, during the sparsification step, we set $w_{L^p} = 0.0002$.

\bigskip

\begin{table}[hbt]
    \centering 
    \rowcolors{2}{cyan!10}{white}
    \begin{threeparttable}
        \caption{Experimental results for the Allen-Cahn equation} 
        \label{Table:Experiments:Allen_Cahn}

        \begin{tabulary}{.9\linewidth}{p{0.7cm}p{0.6cm}p{1.3cm}p{1.0cm}p{1.5cm}L}
            \toprule[0.3ex]
            $N_{Data}$ & Noise & $N_{Burn-in}$ & $N_{Sparse}$ & $N_{Fine-tune}$ & Identified PDE \\
            \midrule[0.1ex]
            $4,000$ & $25\%$ & $2,000$ & $1,000$ & $10,000$ & $D_t U =  0.9576(U) +  0.0029(D_x^2 U) -  0.9673(U)^3$ \\
            \addlinespace[0.4em]
            $4,000$ & $50\%$ & $2,000$ & $1,000$ & $10,000$ & $D_t U =  0.9165(U) +  0.0026(D_x^2 U) -  0.9290(U)^3$ \\
            \addlinespace[0.4em]
            $4,000$ & $75\%$ & $2,000$ & $1,000$ & $10,000$ & $D_t U =  0.9391(U) +  0.0028(D_x^2 U) -  0.9123(U)^3$ \\
            \addlinespace[0.4em]
            $4,000$ & $100\%$ & $2,000$ & $1,000$ & $10,000$ & $D_t U =  0.6921(U) +  0.0020(D_x^2 U) -  0.6211(U)^3$ \\
            \addlinespace[0.4em]
            $1,000$ & $25\%$ & $2,000$ & $1,000$ & $10,000$ & $D_t U =  0.8686(U) +  0.0028(D_x^2 U) -  0.8158(U)^3$ \\ 
            \addlinespace[0.4em]
            $1,000$ & $50\%$ & $2,000$ & $1,000$ & $10,000$ & $D_t U =  0.1405(U)  -  0.0910(U)^3$ \\
            \addlinespace[0.4em]
            $500$ & $10\%$ & $2,000$ & $1,000$ & $10,000$ & $D_t U =  0.9473(U) +  0.0028(D_x^2 U) -  0.9386(U)^3$ \\
            \addlinespace[0.4em]
            $500$ & $25\%$\tnote{1} & $2,000$ & $1,000$ & $10,000$ & $D_t U =  0.8562(U) +  0.0025(D_x^2 U) -  0.8239(U)^3$ \\
            \bottomrule[0.3ex]
        \end{tabulary}

        \begin{tablenotes}
            \item [1] For this experiment, we use $W_{L^p} = 0.0003$ during the sparsification step. If we set $w_{L^p} = 0.0002$, \texttt{PDE-LEARN} identifies $D_t U = 0.4520(U) + 0.0018(D_x^2 U) - 0.3729(U)^3 + 0.0019(D_x U)^2(U)$.
        \end{tablenotes}

    \end{threeparttable}
\end{table}

Thus, \texttt{PDE-LEARN} correctly identifies the Allen-Cahn equation in all but one of our experiments.
Significantly, in every experiment, the right-hand side of the identified PDE contains the $U^3$ term with its small coefficient.
These results suggest that \texttt{PDE-LEARN} can reliably identify small coefficients in hidden PDEs, even when they are close to the threshold. 
Notably, \texttt{PDE-LEARN} identifies the Allen-Cahn equation with up to $100\%$ noise and $4,000$ data points.
Further, it correctly identifies the Allen-Cahn equation with $25\%$ noise using just $500$ data points.
With that said, \texttt{PDE-LEARN}’s robustness to noise does decrease with the number of data points. 
In particular, it fails to identify the Allen-Cahn equation in the $1,000$ data points, $50\%$ noise experiment.
In this experiment, the identified PDE is too sparse, though the two RHS terms that \texttt{PDE-LEARN} identifies are present in the Allen-Cahn equation.

\subsection{2D Wave equation}
\label{sec:2D-Wave}

All of the experiments we have considered thus far deal with a single spatial variable ($x$) and a Hidden PDE whose left-hand side is a time derivative.
While PDEs of this form are not uncommon in physics, they do not constitute all possible PDEs of practical interest. 
As stated in section \ref{sec:Methodology}, \texttt{PDE-LEARN} can learn a more general class of  PDEs.
In this subsection, we demonstrate this ability by testing \texttt{PDE-LEARN} on the 2D wave equation. 
This second-order PDE appears in electromagnetism, structural mechanics, etc. 
With two spatial variables, the wave equation takes on the following form: 

$$ D_t^2 u = c^2 \left( D_x^2 u + D_y^2 u \right).$$

Here, $c > 0$ is called the wave speed.
Rearranging the above equation gives

\begin{equation} D_x^2 u = \varepsilon D_t^2 u - D_y^2 u, \label{eq:2D-Wave} \end{equation}

where $\varepsilon = 1/c^2$.
We test \texttt{PDE-LEARN} on this form of the 2D wave equation with $\varepsilon = 1$ and the problem domain $(t, x, y) \in (0, 10] \times [-5, 5] \times [-5, 5]$. 
We use a known solution to generate the noise-free data set.
In particular,

$$u(t, x, y) = -\sin\left(t - x\right) + \exp\left(.05(t - x - y)\right) + \sin\left(t - y\right),$$

satisfies the heat equation on the problem domain.
To make our noise-free data set, we evaluate this function at $4,000$ points drawn from a uniform distribution over the problem domain. 
We then corrupt this data set using varying noise levels, engendering our noisy and limited data sets.

\bigskip 

In all of our experiments with the wave equation, $U$ contains four layers with $40$ neurons per layer.
For these experiments, we use the left-hand side term

$$f_0 \Big(\hat{\partial}^0 u, \ldots, \hat{\partial}^{N_M} u \Big) = D_x^2 U.$$

Further, we use the following library of right-hand side terms 

$$ \begin{aligned} U, D_t U, D_t^2 U, &D_t^3 U, D_y U, D_y^2 U, D_y^3 U, \\
(U)^2, (D_t U)U, (D_t^2 U)U, (D_t U)^2&, (D_y U)U, (D_y^2 U)U, (D_y U)^2, (D_t U)(D_y U), \\
(U)^3, (D_t U)(U)^2, (D_t^2 U)U^2, (D_t U)^2U, &(D_y U)U^2, (D_y^2 U)U^2, (D_y U)^2U, (D_t U)(D_y U)U \end{aligned}$$

\bigskip

Table \ref{Table:Experiments:2D_Wave} reports the results of our wave equation experiments.
In each experiment, the burn-in step lasts for $2,000$ epochs ($N_{Burn-in} = 2,000$), the sparsification step lasts $2,000 - 3,000$ epochs ($2,000 \leq N_{Sparse} \leq 3,000$), and the fine-tuning step lasts for up to $1,000$ epochs ($N_{Fine-tune} \leq 1,000$). 
Further, during the sparsification step, we set $w_{L^p} = 0.0003$.

\bigskip

\begin{table}[hbt]
    \centering 
    \rowcolors{2}{white}{cyan!10}
    \begin{threeparttable}
        \caption{Experimental results for the 2D wave equation} 
        \label{Table:Experiments:2D_Wave}

        \begin{tabulary}{.9\linewidth}{p{0.7cm}p{0.6cm}p{1.3cm}p{1.0cm}p{1.5cm}L}
            \toprule[0.3ex]
            $N_{Data}$ & Noise & $N_{Burn-in}$ & $N_{Sparse}$ & $N_{Fine-tune}$ & Identified PDE \\
            \midrule[0.1ex]
            $4,000$ & $0\%$ & $2,000$ & $2,000$ & $1,000$ & $D_x^2 U =  0.9973(D_t^2 U) -  0.9938(D_y^2 U)$ \\
            \addlinespace[0.4em]
            $4,000$ & $25\%$ & $2,000$ & $2,000$ & $1,000$ & $D_x^2 U =  0.9981(D_t^2 U) -  0.9919(D_y^2 U)$ \\
            \addlinespace[0.4em]
            $4,000$ & $50\%$ & $2,000$ & $2,000$ & $1,000$ & $D_x^2 U =  0.9656(D_t^2 U) -  0.9568(D_y^2 U)$ \\
            \addlinespace[0.4em]
            $4,000$ & $75\%$ & $2,000$ & $2,000$ & $400$ & $D_x^2 U =  0.9268(D_t^2 U) -  0.9179(D_y^2 U)$ \\
            \addlinespace[0.4em]
            $4,000$ & $100\%$\tnote{1} & $2,000$ & $3,000$ & $1,000$ & $D_x^2 U =  0.9004(D_t^2 U) -  0.8794(D_y^2 U)$ \\
            \bottomrule[0.3ex]
        \end{tabulary}

        \begin{tablenotes}
            \item [1] For this experiment, we use $W_{L^p} = 0.001$ during the sparsification step. If we set $w_{L^p} = 0.0003$, \texttt{PDE-LEARN} identifies $D_x^2 U =  0.8270(D_t^2 U) -  0.6916(D_y^2 U) -  0.2471(D_y^2 U)(U) -  0.0075(D_t^2 U)(U)^2 +  0.0739(D_y^2 U)(U)^2$.
        \end{tablenotes}

    \end{threeparttable}
\end{table}

Thus, \texttt{PDE-LEARN} successfully identifies the wave equation in all experiments, though we did have to increase $w_{L^p}$ in the $100\%$ noise experiment.
Further, in most cases, the coefficients in the identified PDE closely match those of the true PDE, deviating less than $1\%$ from their true values in the $0$\% and $25$\% noise experiments.
However, as with other equations, these experiments reveal that \texttt{PDE-LEARN} does have some limitations.
If we do not increase $w_{L^p}$ to $0.001$ in the $100$\% noise experiment, the identified PDE contains terms that are not present in the heat equation.
However, even in this case, the identified PDE contains both RHS terms of the wave equation.
Further, the coefficients of the extra terms are significantly smaller than those of the RHS terms present in the wave equation.
This result adds to our observation that \texttt{PDE-LEARN} extracts useful information about the hidden PDE even when it fails.
These experiments demonstrate that \texttt{PDE-LEARN} can identify PDEs with multiple spatial variables and works with arbitrary left-hand side terms.
\section{Discussion} \label{sec:Discussion}

This section discusses further aspects of \texttt{PDE-LEARN}, with a special emphasis on our rationale behind the algorithm's design.
First, in section \ref{sub_sec:Discussion:Hyperparameter}, we discuss hyperparameter selection with \texttt{PDE-LEARN}.
Section \ref{sub_sec:Discussion:Pruning} discusses why pruning (between the burn-in, sparsification, and fine-tuning steps) is necessary.
In section \ref{sub_sec:Discussion:Lp_Loss}, we discuss the $L^p$ loss.
Section \ref{sub_sec:Discussion:Small_Coeff} analyzes why the coefficients in the identified PDE tend to be smaller than the corresponding coefficients in the hidden PDE.
Finally, in section \ref{sub_sec:Discussion:Limits}, we discuss some limitations of \texttt{PDE-LEARN} as well as potential future directions.

\subsection{Hyperparameter Selection}
\label{sub_sec:Discussion:Hyperparameter}

\texttt{PDE-LEARN} contains many hyperparameters, including the library terms, loss function weights, $p$, and network architecture. 
We did not perform hyperparameter selection in our experiments.
Therefore, we do not claim our choices in the experiments are optimal; they may not be suitable in every situation.

\bigskip 

We found that $p = 0.1$ works well for the equations in our experiments.
With that said, other values of $p$ may work well in other situations.
$p$ is a hyperparameter and should be treated as such (with $p = 0.1$ as a good default value).
Our loss function weights work well in our experiments.
However, we believe it may make sense to use different weights for the data and collocation losses if the data set is unusually limited or noisy.
As for the library terms, our experiments suggest that \texttt{PDE-LEARN} can identify sparse PDEs even from a relatively large library of RHS terms. 
Using a large library increases the likelihood that the terms in the hidden PDE are in the library.
Therefore, choosing a large library is a good default choice. 
Finally, for the network architecture, we believe the default choice of network architecture should contain the fewest parameters possible to learn the underlying data set (which can be empirically determined).

\bigskip

\subsection{Pruning} 
\label{sub_sec:Discussion:Pruning}

Even after the sparsification step, most components of $\xi$ are small (a few orders of magnitude above machine epsilon) but non-zero.
One likely reason for this is that \texttt{PDE-LEARN} works with finite precision floating-point arithmetic.
Let $\varepsilon$ denote machine-epsilon.
If a component of $\xi$, $\xi_k$, is smaller than $\sqrt{\varepsilon}$, then $\xi_k^2$ in equation \ref{eq:Loss:Lp} is smaller than machine epsilon, meaning that \texttt{PDE-LEARN} can not accurately compute $\xi^2$.
These results make thresholding necessary and are why we set the threshold slightly above $\sqrt \varepsilon$.

\bigskip 

The biggest drawback of pruning is that \texttt{PDE-LEARN} can not identify coefficients in the hidden PDE whose magnitude is smaller than the threshold. 
In our experiments, we use single-precision floating-point arithmetic, meaning that our threshold is around $10^{-4}$.
However, \texttt{PDE-LEARN} can be implemented using double-precision floating-point arithmetic, allowing for a smaller threshold should the need arise.

\bigskip

\subsection{L\textsuperscript{p} loss}
\label{sub_sec:Discussion:Lp_Loss}

As discussed in section \ref{sec:Methodology}, the $L^p$ loss effectively embeds the Iteratively Reweighted Least Squares loss into our loss function.
At the start of each epoch, the $L^p$ loss is equal to $\| \xi \|_p^p$.
Crucially, however, since the weights, $a_k$, in the $L^p$ loss are treated as constants during back-propagation, the $L^p$ loss is a convex function of $\xi$. 
By contrast, the p-norm $\xi \to \| \xi \|_p^p$ for $0 < p < 1$ is not convex.
In particular, it contains sharp cusps along the coordinate axes.
These cusps make it nearly impossible for standard optimizers (such as the Adam optimizer we use to train \texttt{PDE-LEARN}) to converge to a minimum of the p-norm. To illustrate this point, we tried replacing the $L^p$ loss with $\| \xi \|_p^p$.
Unsurprisingly, this change renders \texttt{PDE-LEARN} unusable; it fails to converge, even when training on noise-free data sets.
Thus, the Iteratively Reweighted Least Squares loss function is an essential aspect of \texttt{PDE-LEARN}.

\bigskip

\subsection{The coefficients in the identified PDEs are too small}
\label{sub_sec:Discussion:Small_Coeff}

In many of our experiments, the coefficients in the identified PDE are smaller than those in the hidden PDE.
This effect appears to get worse as the noise level increases.
We believe this is a result of how we sparsify the PDE. 

\bigskip

During the sparsification step, the $L^p$ loss pushes the components of $\xi$ to zero.
Our choice of $p = 0.1$ means that the $L^p$ loss does a reasonable job of penalizing the number of non-zero terms, though it still pushes all components of $\xi$ towards zero.
In principle, the collocation loss will grow if the components of $\xi$ deviate too much from corresponding values in the hidden PDE.
Thus, the components of $\xi$ must balance the collocation and $L^p$ losses.
The result is usually a compromise; the components of $\xi$ become as small as they can be without causing a significant increase in the collocation loss.
Thus, the components of $\xi$ corresponding to terms in the hidden PDE generally survive the sparsification step but end up with artificially small magnitudes. 
This result is why we include the fine-tuning step, during which the coefficients recover and approach the corresponding values in the hidden PDE. 
With that said, noise makes it difficult for \texttt{PDE-LEARN} to precisely resolve the coefficients.
This means that the collocation loss does not significantly decrease once the components of $\xi$ are reasonably close to the corresponding values in the hidden PDE.
This result may explain why the coefficients in the identified PDE tend to shrink as the noise level increase.

\bigskip

Running more fine-tuning epochs generally improves the accuracy of the identified coefficients. 
However, if the noise is high and the data is limited, the networks can over-fit the data set. 
Over-fitting begins when the testing data loss increases while the training data loss decreases. 
In our experiments, we use an early stopping procedure to stop the fine-tuning step as soon as over-fitting begins.

\bigskip

\subsection{Limitations and Future Directions}
\label{sub_sec:Discussion:Limits}

Our experiments in section \ref{sec:Experiments} demonstrate that \texttt{PDE-LEARN} can learn a wide variety of PDEs.
\texttt{PDE-LEARN} places relatively weak assumptions on the form of the underlying PDE.
While the previous works discussed in section \ref{sec:Related_Work} assume the left-hand side of the hidden PDE is a time derivative, \texttt{PDE-LEARN} can learn PDEs with arbitrary left-hand side terms.
With that said, the hidden PDE must be in the form of equation \ref{eq:PDE}.
Therefore, the user must select an appropriate library of terms. 
Without specialized domain knowledge, selecting an appropriate library may be challenging.

\bigskip

Our implementation of \texttt{PDE-LEARN} exclusively uses monomial library terms. 
However, assuming that monomial terms will work in all cases is not reasonable; thus we point out that our implementation of \texttt{PDE-LEARN} can be modified to work with other library terms.
Even with this change, it is unclear how \texttt{PDE-LEARN} would perform when trying to identify PDEs whose terms are not monomials of the $u_i$'s, and their associated partial derivatives.
Therefore, generalizing \texttt{PDE-LEARN} to use other library terms, and exploring how this impacts \texttt{PDE-LEARN}'s performance, represents a potential future area of research.

\bigskip 

Finally, it is worth noting that \texttt{PDE-LEARN}'s performance varies by equation.
\texttt{PDE-LEARN} had little trouble discovering the Brugers', KdV, Allen-Cahn, and Heat equations.
However, \texttt{PDE-LEARN} only identify the KS equation with $15\%$ noise. 
It is not immediately clear why this equation is more challenging for \texttt{PDE-LEARN} to identify, though past works have reported similar results \cite{chen2021physics}. 
Identifying factors (both in the hidden PDE and the data set) that impact \texttt{PDE-LEARN}'s performance represents an important area of future research. 


\section{Conclusion}
\label{sec:Conclusion}

This paper introduced \texttt{PDE-LEARN}, a novel PDE-discovery algorithm to identify human-readable PDEs from noisy and limited data.
\texttt{PDE-LEARN} utilizes Rational Neural Networks, targeted collocation points, and a three-part loss function inspired by Iteratively Reweighted Least Squares.
Further, unlike many previous works, \texttt{PDE-LEARN} can identify PDEs with multiple spatial variables and arbitrary left-hand side terms (see equation \ref{eq:PDE}). 
The general form of the hidden PDE, equation \ref{eq:PDE}, gives \texttt{PDE-LEARN} tremendous flexibility in discovering PDEs from data.
Our experiments in section \ref{sec:Experiments} demonstrate that \texttt{PDE-LEARN} can identify a variety of PDEs from noisy, limited data sets.

\bigskip 

\texttt{PDE-LEARN} appears to be an effective tool for PDE discovery.
Its ability to identify a variety of linear and non-linear PDEs, including those with multiple spatial variables, suggests that \texttt{PDE-LEARN} may be useful in discovering governing equations for physical systems that, until now, have evaded such descriptions.

\section{Acknowledgements}
This work is supported by the Office of Naval Research (ONR), under grant N00014-22-1-2055.
Further, Robert Stephany is supported his NDSEG fellowship.

\printbibliography

@article{abreu2019mortality,
  title={Mortality causes universal changes in microbial community composition},
  author={Abreu, Clare I and Friedman, Jonathan and Woltz, Vilhelm L Andersen and Gore, Jeff},
  journal={Nature communications},
  volume={10},
  number={1},
  pages={1--9},
  year={2019},
  publisher={Nature Publishing Group}
}

@article{amor2020transient,
  title={Transient invaders can induce shifts between alternative stable states of microbial communities},
  author={Amor, Daniel R and Ratzke, Christoph and Gore, Jeff},
  journal={Science advances},
  volume={6},
  number={8},
  pages={eaay8676},
  year={2020},
  publisher={American Association for the Advancement of Science}
}

@article{baydin2018automatic,
  title={Automatic differentiation in machine learning: a survey},
  author={Baydin, Atilim Gunes and Pearlmutter, Barak A and Radul, Alexey Andreyevich and Siskind, Jeffrey Mark},
  journal={Journal of machine learning research},
  volume={18},
  year={2018},
  publisher={Journal of Machine Learning Research}
}

@inproceedings{chartrand2008iteratively,
  title={Iteratively reweighted algorithms for compressive sensing},
  author={Chartrand, Rick and Yin, Wotao},
  booktitle={2008 IEEE international conference on acoustics, speech and signal processing},
  pages={3869--3872},
  year={2008},
  organization={IEEE}
}

@article{boulle2020rational,
  title={Rational neural networks},
  author={Boull{\'e}, Nicolas and Nakatsukasa, Yuji and Townsend, Alex},
  journal={Advances in Neural Information Processing Systems},
  volume={33},
  pages={14243--14253},
  year={2020}
}

@article{kingma2014adam,
  title={Adam: A method for stochastic optimization},
  author={Kingma, Diederik P and Ba, Jimmy},
  journal={arXiv preprint arXiv:1412.6980},
  year={2014}
}

@article{liu1989limited,
  title={On the limited memory BFGS method for large scale optimization},
  author={Liu, Dong C and Nocedal, Jorge},
  journal={Mathematical programming},
  volume={45},
  number={1},
  pages={503--528},
  year={1989},
  publisher={Springer}
}

@article{basdevant1986spectral,
  title={Spectral and finite difference solutions of the Burgers equation},
  author={Basdevant, Cea and Deville, M and Haldenwang, P and Lacroix, JM and Ouazzani, J and Peyret, R and Orlandi, Paolo and Patera, AT},
  journal={Computers \& fluids},
  volume={14},
  number={1},
  pages={23--41},
  year={1986},
  publisher={Elsevier}
}

@misc{driscoll2014chebfun,
  title={Chebfun guide},
  author={Driscoll, Tobin A and Hale, Nicholas and Trefethen, Lloyd N},
  year={2014},
  publisher={Pafnuty Publications, Oxford}
}

@article{korteweg1895xli,
  title={XLI. On the change of form of long waves advancing in a rectangular canal, and on a new type of long stationary waves},
  author={Korteweg, Diederik Johannes and De Vries, Gustav},
  journal={The London, Edinburgh, and Dublin Philosophical Magazine and Journal of Science},
  volume={39},
  number={240},
  pages={422--443},
  year={1895},
  publisher={Taylor \& Francis}
}

@article{bateman1915some,
  title={Some recent researches on the motion of fluids},
  author={Bateman, Harry},
  journal={Monthly Weather Review},
  volume={43},
  number={4},
  pages={163--170},
  year={1915}
}

@article{rudy2017data,
  title={Data-driven discovery of partial differential equations},
  author={Rudy, Samuel H and Brunton, Steven L and Proctor, Joshua L and Kutz, J Nathan},
  journal={Science Advances},
  volume={3},
  number={4},
  pages={e1602614},
  year={2017},
  publisher={American Association for the Advancement of Science}
}

@article{brunton2016discovering,
  title={Discovering governing equations from data by sparse identification of nonlinear dynamical systems},
  author={Brunton, Steven L and Proctor, Joshua L and Kutz, J Nathan},
  journal={Proceedings of the national academy of sciences},
  volume={113},
  number={15},
  pages={3932--3937},
  year={2016},
  publisher={National Acad Sciences}
}

@article{schaeffer2017learning,
  title={Learning partial differential equations via data discovery and sparse optimization},
  author={Schaeffer, Hayden},
  journal={Proceedings of the Royal Society A: Mathematical, Physical and Engineering Sciences},
  volume={473},
  number={2197},
  pages={20160446},
  year={2017},
  publisher={The Royal Society Publishing}
}

@article{berg2017neural,
  title={Neural network augmented inverse problems for PDEs},
  author={Berg, Jens and Nystr{\"o}m, Kaj},
  journal={arXiv preprint arXiv:1712.09685},
  year={2017}
}

@article{both2021deepmod,
  title={DeepMoD: Deep learning for Model Discovery in noisy data},
  author={Both, Gert-Jan and Choudhury, Subham and Sens, Pierre and Kusters, Remy},
  journal={Journal of Computational Physics},
  volume={428},
  pages={109985},
  year={2021},
  publisher={Elsevier}
}

@article{chen2021physics,
  title={Physics-informed learning of governing equations from scarce data},
  author={Chen, Zhao and Liu, Yang and Sun, Hao},
  journal={Nature communications},
  volume={12},
  number={1},
  pages={1--13},
  year={2021},
  publisher={Nature Publishing Group}
}

@article{stephany2022pde,
  title={PDE-READ: Human-readable partial differential equation discovery using deep learning},
  author={Stephany, Robert and Earls, Christopher},
  journal={Neural Networks},
  volume={154},
  pages={360--382},
  year={2022},
  publisher={Elsevier}
}

@article{guyon2002gene,
  title={Gene selection for cancer classification using support vector machines},
  author={Guyon, Isabelle and Weston, Jason and Barnhill, Stephen and Vapnik, Vladimir},
  journal={Machine learning},
  volume={46},
  number={1},
  pages={389--422},
  year={2002},
  publisher={Springer}
}

@article{raissi2018deep,
  title={Deep hidden physics models: Deep learning of nonlinear partial differential equations},
  author={Raissi, Maziar},
  journal={The Journal of Machine Learning Research},
  volume={19},
  number={1},
  pages={932--955},
  year={2018},
  publisher={JMLR. org}
}

@article{atkinson2019data,
  title={Data-driven discovery of free-form governing differential equations},
  author={Atkinson, Steven and Subber, Waad and Wang, Liping and Khan, Genghis and Hawi, Philippe and Ghanem, Roger},
  journal={arXiv preprint arXiv:1910.05117},
  year={2019}
}

@article{bonneville2021bayesian,
  title={Bayesian Deep Learning for Partial Differential Equation Parameter Discovery with Sparse and Noisy Data},
  author={Bonneville, Christophe and Earls, Christopher J},
  journal={arXiv preprint arXiv:2108.04085},
  year={2021}
}

@article{gurevich2019robust,
  title={Robust and optimal sparse regression for nonlinear PDE models},
  author={Gurevich, Daniel R and Reinbold, Patrick AK and Grigoriev, Roman O},
  journal={Chaos: An Interdisciplinary Journal of Nonlinear Science},
  volume={29},
  number={10},
  pages={103113},
  year={2019},
  publisher={AIP Publishing LLC}
}

@article{messenger2021weak,
  title={Weak SINDy for partial differential equations},
  author={Messenger, Daniel A and Bortz, David M},
  journal={Journal of Computational Physics},
  volume={443},
  pages={110525},
  year={2021},
  publisher={Elsevier}
}

@article{bongard2007automated,
  title={Automated reverse engineering of nonlinear dynamical systems},
  author={Bongard, Josh and Lipson, Hod},
  journal={Proceedings of the National Academy of Sciences},
  volume={104},
  number={24},
  pages={9943--9948},
  year={2007},
  publisher={National Acad Sciences}
}

@article{schmidt2009distilling,
  title={Distilling free-form natural laws from experimental data},
  author={Schmidt, Michael and Lipson, Hod},
  journal={science},
  volume={324},
  number={5923},
  pages={81--85},
  year={2009},
  publisher={American Association for the Advancement of Science}
}

@article{sivashinsky1977nonlinear,
  title={Nonlinear analysis of hydrodynamic instability in laminar flames—I. Derivation of basic equations},
  author={Sivashinsky, Gregory I},
  journal={Acta astronautica},
  volume={4},
  number={11},
  pages={1177--1206},
  year={1977}
}

@article{kuramoto1976persistent,
  title={Persistent propagation of concentration waves in dissipative media far from thermal equilibrium},
  author={Kuramoto, Yoshiki and Tsuzuki, Toshio},
  journal={Progress of theoretical physics},
  volume={55},
  number={2},
  pages={356--369},
  year={1976},
  publisher={Oxford University Press}
}

@article{papageorgiou1991route,
  title={The route to chaos for the Kuramoto-Sivashinsky equation},
  author={Papageorgiou, Demetrios T and Smyrlis, Yiorgos S},
  journal={Theoretical and Computational Fluid Dynamics},
  volume={3},
  number={1},
  pages={15--42},
  year={1991},
  publisher={Springer}
}

@article{allen1979microscopic,
  title={A microscopic theory for antiphase boundary motion and its application to antiphase domain coarsening},
  author={Allen, Samuel M and Cahn, John W},
  journal={Acta metallurgica},
  volume={27},
  number={6},
  pages={1085--1095},
  year={1979},
  publisher={Elsevier}
}

@inproceedings{glorot2010understanding,
  title={Understanding the difficulty of training deep feedforward neural networks},
  author={Glorot, Xavier and Bengio, Yoshua},
  booktitle={Proceedings of the thirteenth international conference on artificial intelligence and statistics},
  pages={249--256},
  year={2010},
  organization={JMLR Workshop and Conference Proceedings}
}

\end{document}